%% file: main.tex
\documentclass{article}

\usepackage{microtype}
\usepackage{graphicx}
\usepackage{subcaption}
\usepackage{booktabs} 

\usepackage{hyperref}

\usepackage[preprint]{icml2026}

\usepackage{amsmath}
\usepackage{amssymb}
\usepackage{mathtools}
\usepackage{amsthm}

\usepackage[capitalize,noabbrev]{cleveref}

\theoremstyle{plain}
\newtheorem{theorem}{Theorem}[section]
\newtheorem{proposition}[theorem]{Proposition}

\theoremstyle{definition}

\theoremstyle{remark}

\usepackage[textsize=tiny]{todonotes}

\usepackage{xcolor}
\usepackage{adjustbox}
\usepackage{bm}
\usepackage{multirow}

\newcommand{\ours}{Euphonium}

\usepackage{pifont}

\newcommand{\myteaser}{
    \begin{center}
        \begin{minipage}{0.35\textwidth}
            \centering
            \includegraphics[width=\linewidth]{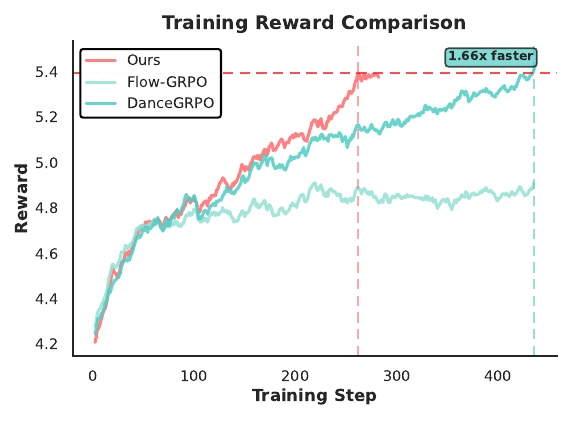}
        \end{minipage}
        \begin{minipage}{0.59\textwidth}
            \centering
            \includegraphics[width=\linewidth]{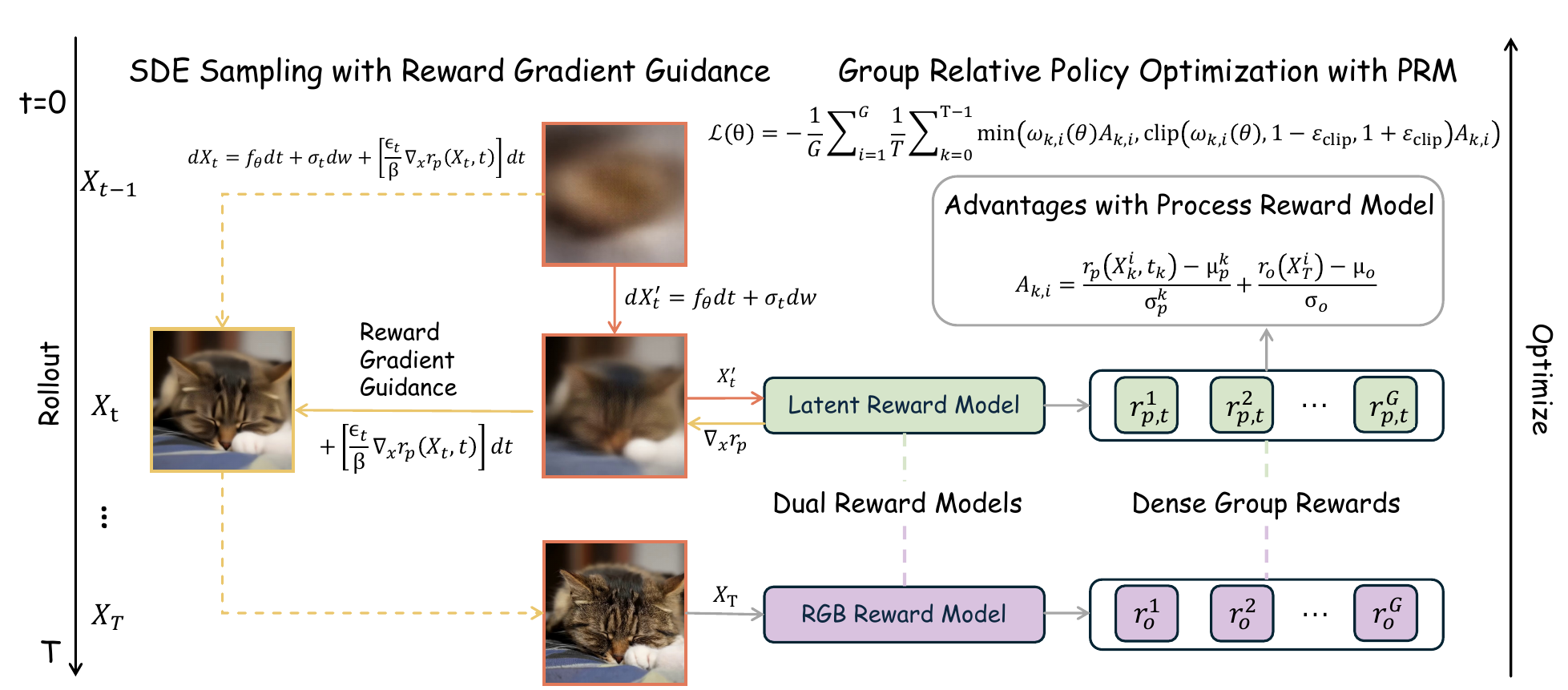}
        \end{minipage}%

        \captionof{figure}{
            \textbf{Left:} Convergence Comparison. {\ours} converges \textbf{1.66}$\times$ faster than other baselines, reaching equivalent performance with fewer training steps.
            \textbf{Right:} Overview of the {\ours} pipeline.
        }
        \label{fig:teaser}
    \end{center}
}

\icmltitlerunning{{\ours}: Steering Video Flow Matching via Process Reward Gradient Guided Stochastic Dynamics}

\begin{document}

\twocolumn[
  \icmltitle{
  \raisebox{-0.20\height}{\includegraphics[height=1.5em]{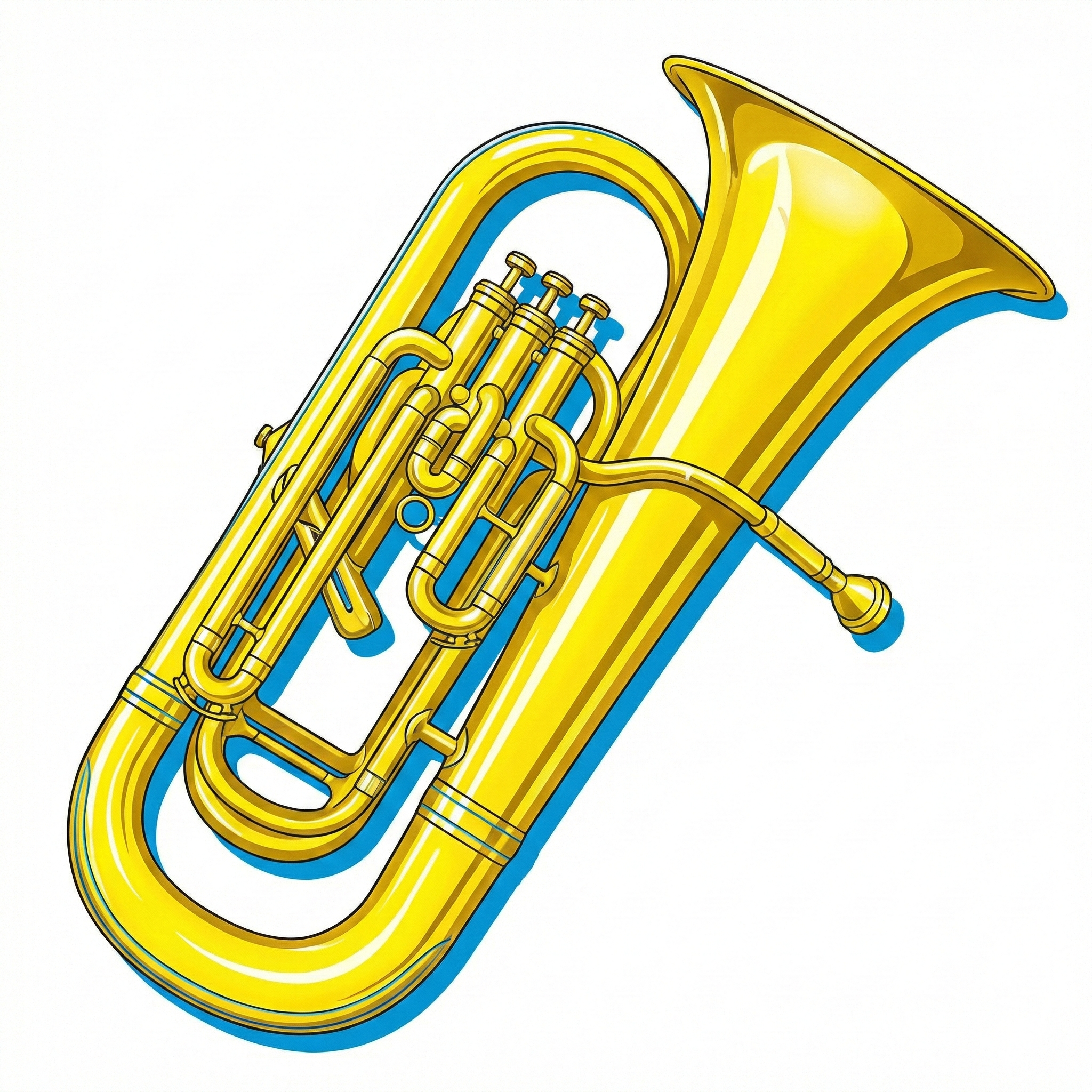}} 
  Euphonium: Steering Video Flow Matching via \\
  Process Reward Gradient Guided Stochastic Dynamics
}



\icmlsetsymbol{equal}{*}
\icmlsetsymbol{intern}{$\dagger$}
\icmlsetsymbol{leader}{$\ddagger$}
\icmlsetsymbol{corresp}{$\S$} 

\begin{icmlauthorlist}
  \icmlauthor{Ruizhe Zhong}{equal,intern,sjtu}
  \icmlauthor{Jiesong Lian}{equal,intern,hust}
  \icmlauthor{Xiaoyue Mi}{ucas}
  \icmlauthor{Zixiang Zhou}{tencent}
  \icmlauthor{Yuan Zhou}{leader,tencent}
  \icmlauthor{Qinglin Lu}{tencent}
  \icmlauthor{Junchi Yan}{corresp,sjtu} 
\end{icmlauthorlist}

\icmlaffiliation{hust}{Huazhong University of Science and Technology}
\icmlaffiliation{sjtu}{Shanghai Jiao Tong University}
\icmlaffiliation{tencent}{Tencent Hunyuan}
\icmlaffiliation{ucas}{University of Chinese Academy of Sciences}

\icmlcorrespondingauthor{Ruizhe Zhong}{zerzerzerz271828@sjtu.edu.cn}
\icmlcorrespondingauthor{Jiesong Lian}{lian700@hust.edu.cn}
\icmlcorrespondingauthor{Junchi Yan}{yanjunchi@sjtu.edu.cn}

\icmlkeywords{Video Generation, Post-Training, Reinforcement Learning, Reward Model}

  \icmlkeywords{Video Generation, Post-Training, Reinforcement Learning, Reward Model}

  \myteaser
  \vskip 0.3in
]




\printAffiliationsAndNotice{%
\icmlEqualContribution 
$^\dagger$Work done during an internship at Tencent Hunyuan 
$^\ddagger$Project Leader
}

\input{section/0-abstract}
\input{section/1-introduction}
\input{section/2-related_works}
\input{section/3-preliminary}
\input{section/4-method}
\input{section/5-experiments}
\input{section/6-conclusion}

\clearpage
\section*{Impact Statement}
This paper presents work whose goal is to advance the field of Machine
Learning. There are many potential societal consequences of our work, none
which we feel must be specifically highlighted here.

\bibliography{ref}
\bibliographystyle{icml2026}

\clearpage
\appendix
\onecolumn
\input{appendix/appendix_intro}
\input{appendix/proof_score_velocity}

\input{appendix/equivalence_flow_grpo}
\input{appendix/optimal_policy}
\input{appendix/latent_prm_justification}
\input{appendix/logprob_derivation}
\input{appendix/implementation_details}
\end{document}

%% file: section/0-abstract.tex
\begin{abstract}
While online Reinforcement Learning has emerged as a crucial technique for aligning flow matching models with human preferences, current approaches are hindered by inefficient exploration during training rollouts.
Relying on undirected stochasticity and sparse outcome rewards, these methods struggle to discover high-reward samples, resulting in data-inefficient and slow optimization.
To address these limitations, we propose {\ours}, a novel framework that steers generation via process reward gradient guided dynamics.
Our key insight is to formulate the sampling process as a theoretically principled Stochastic Differential Equation that explicitly incorporates the gradient of a Process Reward Model into the flow drift.
This design enables dense, step-by-step steering toward high-reward regions, advancing beyond the unguided exploration in prior works, and theoretically encompasses existing sampling methods (e.g., Flow-GRPO, DanceGRPO) as special cases.
We further derive a distillation objective that internalizes the guidance signal into the flow network, eliminating inference-time dependency on the reward model.
We instantiate this framework with a Dual-Reward Group Relative Policy Optimization algorithm, combining latent process rewards for efficient credit assignment with pixel-level outcome rewards for final visual fidelity.
Experiments on text-to-video generation show that {\ours} achieves better alignment compared to existing methods while accelerating training convergence by 1.66$\times$.
Our code is available at \url{https://github.com/zerzerzerz/Euphonium}
\end{abstract}

%% file: section/1-introduction.tex
\section{Introduction}
\label{sec:intro}

Flow matching~\cite{lipman2022flow, liu2022flow, tong2023improving} has emerged as the dominant paradigm for high-fidelity video generation, underpinning leading systems such as Sora~\cite{sora2025sora2}, Veo~\cite{veo2025veo3}, Wan~\cite{wan2025wan}, Kling~\cite{kling2025kling}, and Seedance~\cite{gao2025seedance}.
While large-scale pre-training endows foundation models with remarkable generative capabilities, it is often insufficient for ensuring alignment with nuanced human preferences, such as specific aesthetic styles and prompt adherence.
Consequently, post-training via Reinforcement Learning (RL) has become a critical frontier for bridging this alignment gap~\cite{black2023training}.

However, adapting online RL algorithms (e.g., Group Relative Policy
Optimization, GRPO) to flow matching requires introducing stochasticity into the deterministic probability flow Ordinary Differential Equations (ODEs). Recent approaches like Flow-GRPO~\cite{liu2025flow} and DanceGRPO~\cite{xue2025dancegrpo} address this by formulating Stochastic Differential Equations (SDEs) to enable policy exploration.
Despite this progress, these methods rely on \textit{undirected} stochastic dynamics: the model explores the latent space via random perturbations and receives feedback only after the entire video is generated.
Consequently, discovering high-reward trajectories becomes exceedingly difficult, as random noise rarely steers generation toward the narrow manifold of high-quality samples.
This inefficiency in exploration leads to sparse positive supervision, resulting in data-inefficient and slow optimization. 

We posit that efficient alignment necessitates not merely stochastic exploration, but \textit{guided} exploration. Rather than relying on undirected noise to fortuitously discover high-reward trajectories, the sampling process itself should be actively steered toward preferred regions. 
To this end, we propose \textbf{{\ours}} (\underline{E}xploration \underline{U}tilizing \underline{P}rocess \underline{H}euristics \underline{O}ver \underline{N}on-equilibr\underline{ium} Sampling), a principled framework grounded in the Non-Equilibrium Transport Sampling (NETS) formalism~\cite{albergo2023stochastic, albergo2024nets}. 
By defining a novel reward-augmented potential that integrates the flow prior with a dense Process Reward Model (PRM)~\cite{lightman2023let}, we derive a theoretically grounded SDE where the reward gradient is explicitly incorporated into the flow drift.
This design enables dense, step-by-step steering toward high-reward regions during exploration.
Moreover, our formulation theoretically encompasses existing sampling methods (e.g., Flow-GRPO, DanceGRPO) as special cases when the reward term vanishes, providing a unified perspective on stochastic flow dynamics.

A potential concern with reward-guided exploration is the inference-time dependency on the reward model. We address this by deriving a distillation objective that internalizes the guidance signal into the flow network weights, enabling the deployed model to operate identically to the base generator without requiring an external reward model.

We instantiate this framework via a Dual-Reward GRPO algorithm, combining latent-space process rewards for efficient credit assignment with pixel-space outcome rewards for final visual fidelity alignment.
Our main contributions are summarized as follows:
\begin{itemize}
    \item \textbf{Guided Exploration via Process Reward Gradient.} 
    We derive a reward-augmented stochastic dynamics that injects dense gradient signals from a Process Reward Model directly into the flow drift. 
    This principled formulation enables active steering toward high-reward regions during exploration, and theoretically encompasses prior sampling strategies (e.g., Flow-GRPO, DanceGRPO) as reward-free special cases.
    
    \item \textbf{Dual-Reward Optimization.} 
    We introduce a dual-reward training scheme that combines latent-space process rewards for efficient credit assignment with pixel-space outcome rewards for final visual fidelity alignment.
    
    \item \textbf{Reward-Gradient-Free Inference.} 
    To eliminate inference-time dependency on the reward model after training, we derive a distillation objective that internalizes the guidance signal into the flow network weights, enabling deployment identical to the base generator.
    
    \item \textbf{Empirical Performance.} 
    {\ours} achieves better alignment on VBench2 for text-to-video generation, outperforming existing post-training methods while accelerating training convergence by \textbf{1.66}$\times$.
\end{itemize}

%% file: section/2-related_works.tex
\section{Related Works}

\subsection{Flow Matching}
Flow matching~\cite{lipman2022flow, tong2023improving} has become one of the standards for video generation. Specifically, Rectified Flow~\cite{liu2022flow, esser2024scaling} utilizes linear optimal transport (OT) paths to enforce straight trajectories, reducing inference steps.  

\subsection{Flow Model Alignment}
To align generative models with human preferences, RL techniques developed for diffusion models~\cite{black2023training} have been adapted to flow matching. Existing methods generally fall into two paradigms: stochastic exploration and direct optimization.

\textbf{Stochastic Exploration.} 
A key challenge in applying RL to flow matching is the deterministic nature of the ODE sampler. Flow-GRPO~\cite{liu2025flow} addresses this by formulating a SDE to inject noise, enabling policy exploration around the probability flow. DanceGRPO~\cite{xue2025dancegrpo} improves this with a shared-noise strategy to isolate policy-driven improvements, while MixGRPO~\cite{li2025mixgrpo} adopts a mixed ODE-SDE strategy to reduce computational overhead. To mitigate the sparsity of outcome rewards, Chunk-GRPO~\cite{luo2025sample} and E-GRPO~\cite{zhang2026grpo} propose step aggregation, while TempFlow-GRPO~\cite{he2025tempflow} and TreeGRPO~\cite{ding2025treegrpo} leverage branching structures for finer credit assignment. Despite these advances, these methods rely on undirected noise, exploring the latent space inefficiently without active guidance.

\textbf{Direct Optimization.}
Distinct from stochastic RL, another paradigm utilizes reward gradients directly. ReFL~\cite{xu2023imagereward} performs direct backpropagation from the reward model through the denoising chain to update parameters. Furthermore, it requires ground truth samples for supervised regularization. VGG-Flow~\cite{liu2025value} adopts an optimal control perspective, regressing the velocity field to match the gradient of a learned value function. 
However, these methods often rely on deterministic regression that limits diversity and exploration.

\subsection{Guided Sampling and Process Supervision}
Generative processes can also be steered via inference-time mechanisms. Early works like Classifier Guidance~\cite{ho2022classifier} and energy-based guidance~\cite{lu2023contrastive} modify the drift term based on conditioning. Process supervision, popularized in LLMs~\cite{lightman2023let}, has been adapted to video generation. Video-T1~\cite{liu2025video} applies search-based methods using verifiers to prune low-quality branches. 
While VideoAlign~\cite{liu2025improving} pioneered the utilization of video generation models as latent reward models, its application was restricted to inference-time guidance. 
Building upon this latent supervision concept, PRFL~\cite{mi2025video} incorporates a process-aware latent reward model to optimize video generation during training via direct reward feedback learning. However, PRFL employs a deterministic sampling strategy and requires ground truth samples for supervised regularization.

%% file: section/3-preliminary.tex
\section{Preliminaries}
\label{sec:prelim}
\subsection{Flow Matching and Linear OT Paths}

Let $p_{\text{data}}(x)$ denote the data distribution and $p_0(x) = \mathcal{N}(x; 0, I)$ be a standard Gaussian source distribution. Continuous Normalizing Flows~\cite{lipman2022flow} define a probability path $p_t(x)$ for $t \in [0, 1]$ generated by a time-dependent vector field. We denote the optimal vector field as $u_t(x)$ and its neural network approximation as $u_\theta(x, t)$ with parameters $\theta$.
The flow $\phi_t(x)$ satisfies the ODE:
\begin{equation}
    \frac{d}{dt}\phi_t(x) = u_t(\phi_t(x)), \quad \phi_0(x) = x.
\end{equation}

Flow matching optimizes $u_\theta(x, t)$ to approximate the conditional vector field generating a target probability path. We adopt the Linear Optimal Transport (OT) conditional path:
\begin{equation}
\label{eq:linear_ot_path}
    X_t = (1-t)X_0 + tX_1,
\end{equation}
where $X_0 \sim p_0$ and $X_1 \sim p_{\text{data}}$. Under this path construction, the marginal score function $\nabla \log p_t(x)$ admits an explicit relationship with the velocity field $u_t(x)$.

\begin{proposition}[Score-Velocity under Linear OT~\cite{zheng2023guided}]
\label{prop:score_velocity}
For the linear path $X_t = (1-t)X_0 + tX_1$ where $X_0 \sim \mathcal{N}(0, I)$, the score function of the marginal density $p_t(x)$ relates to the optimal velocity field $u_t(x)$ via:
\begin{equation} \label{eq:score_explicit}
    \nabla \log p_t(x) = - \frac{x - t u_t(x)}{1-t}.
\end{equation}
\end{proposition}
Detailed proof is provided in Appendix~\ref{app:proof_score_velocity}.

\subsection{Non-Equilibrium Transport Sampling (NETS)}
\label{sec:nets_general}

To enable step-wise guidance during the generation process, we formulate the sampling problem as sampling from a target density $q_t(x)$ defined by a Boltzmann-Gibbs distribution associated with a time-dependent potential $U_t(x)$~\cite{albergo2024nets}:
\begin{equation}
\label{eq:target_density_wrt_potential}
    q_t(x) \propto \exp\left( -U_t(x) \right).
\end{equation}
To sample from $q_t(x)$ while respecting the transport dynamics of the pre-trained flow, we utilize the Non-Equilibrium Transport Sampler (NETS)~\cite{albergo2024nets}. The dynamics are governed by the following SDE:
\begin{equation} \label{eq:general_sde}
    dX_t = \left( u_\theta(X_t, t) - \epsilon_t \nabla U_t(X_t) \right) dt + \sqrt{2\epsilon_t} dW_t,
\end{equation}
where $u_\theta(X_t, t)$ provides the base velocity, $-\nabla U_t$ acts as a conservative force guiding the particle towards low-potential regions, and $\epsilon_t$ is a time-dependent diffusion coefficient. 

%% file: section/4-method.tex
\section{Methodology}
\label{sec:method}
In this section, we present our post-training algorithm for video generation grounded in online reinforcement learning. Our framework enhances the standard sampling process by introducing a PRM that guides exploration through the stochastic dynamics of \cref{sec:prelim}. This guided exploration is integrated with dual-reward GRPO to align the flow matching model with human preferences. 
We first instantiate the framework by deriving guided dynamics for two potential structures (\cref{sec:guided_dynamics}).
We then detail Process Reward Model training (\cref{sec:prm_training}) and the complete online sampling and optimization pipeline (\cref{sec:pipeline}).
Finally, we introduce a policy distillation formulation that explicitly internalizes reward guidance into the flow network, enabling efficient reward-gradient-free inference (\cref{sec:distillation}).

\subsection{Guided Dynamics for Specific Potentials}
\label{sec:guided_dynamics}

We now derive the update rules for distinct sampling scenarios by defining the specific structure of the potential $U_t(x)$ within the general SDE defined in \cref{eq:general_sde}.

\subsubsection{Unguided Stochastic Sampling}
We first consider the case where the potential corresponds to the negative log-likelihood of the flow density:
\begin{equation}
    U_t(x) = -\log p_t(x).
\end{equation}
Substituting into \cref{eq:general_sde} yields the standard score-based diffusion term $\epsilon_t \nabla \log p_t(x)$. Invoking Proposition~\ref{prop:score_velocity} and substituting the velocity $u_\theta$ gives:
\begin{equation} \label{eq:sde_density_explicit}
    dX_t = \left[ \left(1 + \frac{t \epsilon_t}{1-t}\right)u_\theta(X_t, t) - \frac{\epsilon_t}{1-t}X_t \right] dt + \sqrt{2\epsilon_t} dW_t.
\end{equation}
\textbf{Remark.} This formulation recovers the intrinsic stochastic dynamics of the flow model. Without external reward signals, our derivation naturally recovers the dynamics underlying methods such as DanceGRPO and Flow-GRPO. More details are provided in Appendix~\ref{app:equivalence_flow_grpo}.

\subsubsection{Regularized Reward Guidance}
\label{sec:regularized_reward_guidance}
We adopt a more general setting: maximizing expected reward subject to a Kullback-Leibler (KL) divergence constraint against the reference policy $\pi_{\text{ref}}$. The objective is to find a policy $\pi$ maximizing:
\begin{equation}
    \mathcal{J}(\pi) = \mathbb{E}_{x \sim \pi}[r(x)] - \beta D_{\text{KL}}(\pi \| \pi_{\text{ref}}),
\end{equation}
where $\beta$ controls the KL penalty strength. The closed-form solution is the Boltzmann distribution (Appendix~\ref{app:optimal_policy}):
\begin{equation} \label{eq:optimal_policy}
    \pi^*(x) = \frac{1}{Z} \pi_{\text{ref}}(x) \exp\left( \frac{r(x)}{\beta} \right),
\end{equation}
where $Z$ is the partition function.

To apply this result to our time-dependent generative process, we instantiate \cref{eq:optimal_policy} at each timestep $t$. Concretely, we identify the reference policy $\pi_{\text{ref}}$ with the flow marginal $p_t(x)$ and the reward $r(x)$ with the PRM $r_p(x, t)$. The target density $q_t(x)$ then takes the form:
\begin{equation}
    q_t(x) \propto p_t(x) \exp\left( \frac{r_p(x, t)}{\beta} \right).
\end{equation}
According to \cref{eq:target_density_wrt_potential}, we obtain:
\begin{equation} \label{eq:final_potential}
    U_t(x) = -\log p_t(x) - \frac{1}{\beta} r_p(x, t) + C_t,
\end{equation}
where $C_t$ is a constant independent of $x$.

Substituting into \cref{eq:general_sde} and applying Proposition~\ref{prop:score_velocity}, we obtain the KL-regularized dynamics:
\begin{equation} \label{eq:sde_kl_explicit}
\begin{aligned}
    dX_t &= \underbrace{\left[ \left(1 + \frac{t \epsilon_t}{1-t}\right)u_\theta(X_t, t) - \frac{\epsilon_t}{1-t}X_t \right] dt}_{\text{Reference Dynamics}} \\
    &\quad + \underbrace{\left[\frac{\epsilon_t}{\beta} \nabla_x r_p(X_t, t) \right] dt}_{\text{Reward Gradient Guidance}} + \sqrt{2\epsilon_t} dW_t.
\end{aligned}
\end{equation}
Here, $\beta$ governs the KL regularization strength. Higher $\beta$ tightens adherence to the reference flow, while lower $\beta$ permits stronger reward-gradient steering.

\subsection{Training of the Process Reward Model}
\label{sec:prm_training}

To provide dense, step-wise guidance via the potential $U_t(x)$, we train a PRM
$r_\phi(x, t, c)$, where $c$ denotes the text prompt. We adopt a latent-space design to circumvent the prohibitive memory costs of backpropagating through the video decoder and the high variance of zeroth-order gradient estimation (see Appendix~\ref{app:latent_prm_justification} for detailed analysis). The PRM estimates the quality of intermediate latent states $X_t$ throughout the generative trajectory.

\textbf{Architecture.}
The PRM operates directly in the latent space, taking the noisy video latent $x_t$, timestep $t$, and text embeddings $c$ as inputs. We initialize the PRM from a subset of the pre-trained DiT layers to leverage learned representations while reducing computational overhead. A lightweight MLP head projects the hidden states into a scalar reward score $s \in \mathbb{R}$. Both VAE and text encoder are frozen.

\textbf{Noise Perturbation.}
During training, given a clean video latent $X_1 \sim p_{\text{data}}$, we sample a random timestep $t \sim \mathcal{U}[0, 1]$ and construct the perturbed latent $X_t$ following the Linear OT path (\cref{eq:linear_ot_path}). This perturbation strategy ensures that the reward surface $r_\phi$ is well-defined across the entire probability path, enabling reliable gradient estimation at arbitrary noise levels.

\textbf{Training Objective.}
We adopt the Bradley-Terry model~\cite{bradley1952rank} for preference learning. Given a pair of latents $(X_1^w, X_1^l)$ where $X_1^w \succ X_1^l$, we sample a shared timestep $t$ and noise $\epsilon$ to construct the perturbed pair $(X_t^w, X_t^l)$. The model is trained to assign higher reward to the preferred sample via pairwise ranking loss:
\begin{equation}
    \mathcal{L}_{\text{BT}}(\phi) = - \mathbb{E}_{\textstyle \substack{(X_1^w, X_1^l), t, \epsilon}} \left[ \log \sigma \left( r_\phi(X_t^w, t, c) - r_\phi(X_t^l, t, c) \right) \right].
\end{equation}

\subsection{Online Sampling and Training Pipeline}
\label{sec:pipeline}
Our training framework alternates between two phases: (1) Exploration, where the model generates diverse sample trajectories via reward-guided SDE dynamics, and (2) Optimization, where model parameters are updated via GRPO.

\subsubsection{Sampling and Exploration}
Given a prompt $c$ and shared initial noise $X_0 \sim \mathcal{N}(0, I)$, we generate $G$ trajectories $\{X^i_{0:1}\}_{i=1}^G$. Sample diversity arises naturally from the stochastic Wiener process $W_t$ in our SDE formulation (\cref{eq:sde_kl_explicit}). We discretize the continuous-time dynamics via Euler-Maruyama. With step size $\Delta t$, the update rule becomes:
\begin{equation} \label{eq:discretized_sde}
    X^i_{k+1} = X^i_k + \mathcal{D}_{\text{total}}(X^i_k, t_k) \Delta t + \sqrt{2\epsilon_{t_k} \Delta t} Z^i_k,
\end{equation}
where $Z^i_k \sim \mathcal{N}(0, I)$. The drift $\mathcal{D}_{\text{total}}$ combines the flow matching prior with the process reward gradient:
\begin{equation}
    \mathcal{D}_{\text{total}}(X, t) = \left(1 + \frac{t \epsilon_t}{1-t}\right)u_\theta(X, t) - \frac{\epsilon_t}{1-t}X + \frac{\epsilon_t}{\beta} \nabla_x r_p(X, t).
\end{equation}


\subsubsection{Optimization via GRPO}
\textbf{Dual Reward Integration.}
For policy optimization, we employ GRPO with a dual-reward formulation: a step-wise process reward $r_p(x, t)$ and a trajectory-level outcome reward $r_o(x)$.
We integrate these signals by defining the total advantage at step $k$ as the sum of their group-normalized values.
Concretely, for the $i$-th sample:
\begin{equation}
    A_{k,i} = \underbrace{\frac{r_p(X^i_k, t_k) - \mu_p^k}{\sigma_p^k}}_{\text{Process Advantage}} + \underbrace{\frac{r_o(X_{T}^{i}) - \mu_o}{\sigma_o}}_{\text{Outcome Advantage}},
\end{equation}
where $\mu$ and $\sigma$ denote the mean and standard deviation computed over the sample group $G$.

\textbf{Objective Function.}
We define the per-step importance weight between the current policy $\pi_\theta$ and the reference policy $\pi_{\theta_{\text{old}}}$ as $\omega_{k, i}(\theta) = \frac{\pi_\theta(X^i_{k+1} | X^i_k)}{\pi_{\theta_{\text{old}}}(X^i_{k+1} | X^i_k)}$.
The GRPO objective is then formulated using the total advantage $A_{k,i}$:
\begin{equation}
\label{eq:grpo_loss}
\begin{split}
    \mathcal{L}_{\text{GRPO}}(\theta) = -\frac{1}{G} \sum_{i=1}^G \frac{1}{T} \sum_{k=0}^{T-1} \min \Big( & \omega_{k, i}(\theta) A_{k,i}, \\
    \text{clip}(\omega_{k, i}(\theta), 1-\varepsilon_{\text{clip}}, & 1+\varepsilon_{\text{clip}}) A_{k,i} \Big).
\end{split}
\end{equation}
The procedure is summarized in \cref{alg:training_pipeline}. Derivations log-probability used for $\omega_{k, i}(\theta)$ appear in \cref{sec:distillation}.

\input{algorithm/sampling_training_loop}

\subsection{Distillation for Reward-Gradient-Free Inference}
\label{sec:distillation}

The reward gradient $\frac{\epsilon_t}{\beta} \nabla_x r_p$ is applied \textit{exclusively} during the training phase. Once training is complete, the optimized network $u_\theta$ generates samples via standard flow dynamics, \textit{without} computing $\nabla_x r_p$. This design choice is motivated by practical considerations: retaining the reward gradient would necessitate concurrently loading the PRM alongside the video generator, increasing memory footprint and system complexity. However, this raises a natural concern regarding the distribution shift between the guided dynamics used during training and the unguided dynamics used for final inference.

To address this, we introduce a Policy Distillation formulation that explicitly internalizes the reward guidance into the flow network. We treat the reward-guided trajectory generation as a ``teacher'' process and optimize the ``student'' flow network to replicate this guided behavior without explicit reward computation. During the exploration phase of training, trajectory steps $\{X_{k+1}^i\}$ are generated using the full guided dynamics (\cref{eq:discretized_sde}), which incorporates the reward gradient term.
However, when computing the log-probability of the guided steps $X_{k+1}^i$, both the behavior policy $\pi_{\theta_{\text{old}}}$ and the target policy $\pi_\theta$ are evaluated \textit{without} explicit reward guidance in their drifts.
Denoting the policy parameters as $\psi \in \{\theta, \theta_{\text{old}}\}$, we have:
\begin{equation}
    \mu_\psi(X_k, t_k) = X_k + \Delta t \cdot \mathcal{D}_{\text{flow}}(X_k, t_k; \psi),
\end{equation}
where $\mathcal{D}_{\text{flow}}$ denotes the flow-only component of the drift, excluding the reward gradient term. The log-probability of generating the reward-guided step $X_{k+1}$ under policy $\pi_\psi$ is computed as ($C_k$ is a timestep-dependent constant):
\begin{equation}
    \log \pi_\psi(X_{k+1} | X_k) = C_k - \frac{1}{4\epsilon_{t_k} \Delta t} \left\| X_{k+1} - \mu_\psi(X_k, t_k) \right\|^2.
\end{equation}

Since $X_{k+1}$ is generated via reward-guided dynamics while $\mu_\psi$ excludes the reward gradient, the residual $(X_{k+1} - \mu_\psi)$ implicitly encodes the reward signal. Minimizing this residual forces the student velocity field $u_\theta$ to internalize the guidance, enabling the optimized network to steer toward high-reward regions during inference without external reward computation. We provide detailed derivations and empirical comparisons in Appendix~\ref{app:logprob_derivation}.

%% file: algorithm/sampling_training_loop.tex
\begin{algorithm}[tb]
\caption{RGG Sampling \& Distillation Training}
\label{alg:training_pipeline}
\begin{algorithmic}[1]
\STATE {\bfseries Input:} Pre-trained $u_\theta$, PRM $r_p$, ORM $r_o$
\FOR{iteration $= 1, \dots, M$}
    \STATE $\theta_{\text{old}} \gets \theta$
    \STATE \textbf{// Phase 1: Guided Exploration}
    \FOR{each prompt $c, i \in \{1..G\}$}
        \FOR{$k = 0$ to $T-1$}
            \STATE $\mathcal{D}_{\text{flow}} \gets \text{FlowDrift}(X^i_k; \theta_{\text{old}})$
            \STATE $\mathcal{D}_{\text{total}} \gets \mathcal{D}_{\text{flow}} + \frac{\epsilon_{t_k}}{\beta} \nabla_x r_p(X^i_k)$
            \STATE $X^i_{k+1} \gets \text{Step}(X^i_k, \mathcal{D}_{\text{total}})$
            
            \STATE $\mu_{\text{old}} \gets X^i_k + \mathcal{D}_{\text{flow}} \Delta t$ 
            \STATE $\log \pi_{\text{old}} \gets \log \mathcal{N}(X^i_{k+1} | \mu_{\text{old}}, \sigma_t^2)$
            \STATE Store $(X^i_k, X^i_{k+1}, \log \pi_{\text{old}}, r_p(X^i_k))$
        \ENDFOR
        \STATE Compute $A^i_k$ using standardized $r_p, r_o$
    \ENDFOR
    
    \STATE \textbf{// Phase 2: Optimization}
    \FOR{minibatch $b \in \mathcal{B}$}
        \STATE $\mu_{\theta} \gets X_k + \text{FlowDrift}(X_k; \theta) \Delta t$
        \STATE $\log \pi_{\theta} \gets \log \mathcal{N}(X_{k+1} | \mu_{\theta}, \sigma_t^2)$
        \STATE $\omega \gets \exp(\log \pi_{\theta} - \log \pi_{\text{old}})$
        \STATE Update $\theta$ to max $\mathbb{E}[\min(\omega A, \text{clip}(\omega)A)]$
    \ENDFOR
\ENDFOR
\end{algorithmic}
\end{algorithm}

%% file: section/5-experiments.tex

\section{Experiments}
\label{sec:experiments}

\subsection{Experimental Setup}

\textbf{Post-Training Backbone \& Baselines.} 
We empirically validate {\ours} on text-to-video tasks. We implement all methods based on the open-source HunyuanVideo~\cite{kong2024hunyuanvideo} with 14B parameters.
We compare {\ours} against the base model and other leading post-training baselines: Flow-GRPO~\cite{liu2025flow} and DanceGRPO~\cite{xue2025dancegrpo}.

\textbf{Reward Models.}
Both PRM and ORM are trained on a curated dataset emphasizing visual quality and motion fidelity.
For the ORM, we train a scoring model based on InternVL3-1B~\cite{zhu2025internvl3}, optimized using the Bradley-Terry objective on pairwise preference annotations.
For the PRM, we fine-tune a lightweight variant of the HunyuanVideo DiT on the same preference data to predict quality scores directly from latent states $x_t$, conditioned on the text prompt $c$ and diffusion timestep $t$.

\textbf{Datasets.}
To train the reward model, we curate a dataset of 200,000 video samples generated from 20,000 unique prompts. We employ a pointwise binary annotation protocol, labeling samples as positive or negative based on visual quality and motion fidelity. Training pairs are then constructed by contrasting positive and negative instances to enable preference learning. For GRPO training, we collect 10,000 prompts from DanceGRPO~\cite{xue2025dancegrpo} and internal portrait-centric sources, strictly held out from the reward model training set.

\textbf{Evaluation Metrics.}
Performance is evaluated using VBench2~\cite{zheng2025vbench}.
We report the {Total Score} alongside all five sub-scores.

\subsection{Main Results}
\label{sec:main_results}

\begin{figure*}[!tb]
    \centering
    \includegraphics[width=\textwidth]{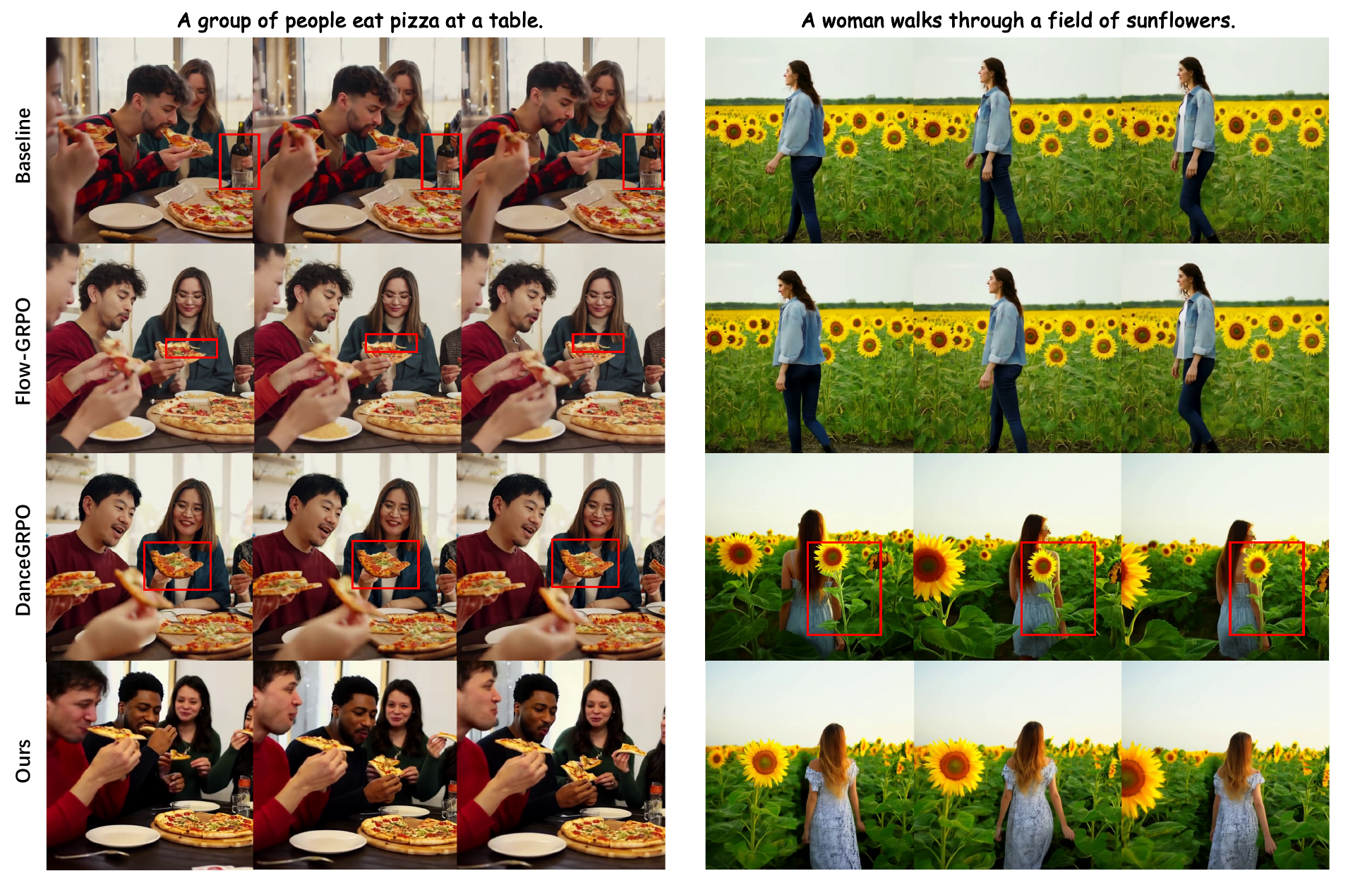}
    \caption{\textbf{Qualitative comparison.} Visual results demonstrate that {\ours} generates more coherent and prompt-aligned videos compared to the pre-trained model, Flow-GRPO and DanceGRPO.}
    \label{fig:visual_comparison}
\end{figure*}

\begin{table*}[!tb]
\caption{\textbf{Quantitative Comparison on VBench2.} The best results are highlighted in \textbf{bold}, and the second-best are \underline{underlined}.}
\label{tab:main_results}
\begin{center}
\begin{adjustbox}{max width=\textwidth}
\begin{tabular}{lcccccc}
\toprule
\textbf{Method} & \textbf{Total Score} $\uparrow$ & \textbf{Creativity} $\uparrow$ & \textbf{Commonsense} $\uparrow$ & \textbf{Controllability} $\uparrow$ & \textbf{Human Fidelity} $\uparrow$ & \textbf{Physics} $\uparrow$ \\
\midrule
Base Model~\cite{kong2024hunyuanvideo} & 51.09 & 40.18 & \underline{62.87} & 23.95 & 83.30 & \underline{45.15} \\
\midrule
Flow-GRPO \cite{liu2025flow} & 51.52 & \textbf{42.42} & 62.86 & 24.41 & 85.41 & 42.48 \\
DanceGRPO \cite{xue2025dancegrpo} & \underline{51.85} & 40.93 & 61.99 & \underline{25.08} & \underline{88.10} & 43.14 \\
\midrule
\textbf{\ours{} (Ours)} & \textbf{54.24} & \underline{41.42} & \textbf{67.17} & \textbf{26.88} & \textbf{88.91} & \textbf{46.84} \\
\bottomrule
\end{tabular}
\end{adjustbox}
\end{center}
\end{table*}

\textbf{Quantitative Analysis.}
\cref{tab:main_results} presents the quantitative performance comparison on VBench2.
{\ours} demonstrates significant improvement, achieving the highest Total Score of 54.24 and outperforming the second-best method, DanceGRPO.
Notably, our method secures the top rank in four out of five dimensions: Commonsense, Controllability, Human Fidelity, and Physics.
While Flow-GRPO marginally leads in Creativity, {\ours} ranks a close second, proving that our approach effectively balances diverse generation with strict adherence to physical constraints and user instructions.

\textbf{Visual Comparison.}
\cref{fig:visual_comparison} visualizes sample frames generated by {\ours} and other methods.
While baselines occasionally exhibit artifacts or fail to capture prompt details, {\ours} generates videos with higher fidelity and improved prompt adherence, corroborating the effectiveness of our approach.

\subsection{Training Efficiency \& Computational Overhead}
\label{sec:efficiency}

A primary consideration for Reward-Gradient Guidance (RGG) is whether the additional per-step computation is justified by improved training efficiency. We analyze both the computational overhead and the convergence behavior.

\textbf{Computational Cost.}
As reported in \cref{tab:rgg_cost}, computing the latent PRM gradient adds only 2.4\% latency per sampling step. The memory overhead is similarly modest at 8.5\%. This efficiency stems from our lightweight PRM design: operating in the compressed latent space with only 8 DiT layers, compared to the 14B-parameter generative backbone with 60 layers. The marginal overhead ensures that RGG remains practical for large-scale video generation training.

\begin{table}[!tb]
\caption{\textbf{RGG Overhead Analysis.} Wall-clock time and peak memory are measured on a single H800 GPU. The lightweight latent-space PRM introduces minimal overhead.}
\label{tab:rgg_cost}
\begin{center}
\begin{adjustbox}{max width=0.8\linewidth}
\begin{tabular}{lccc}
\toprule
\textbf{Metric} & \textbf{Baseline} & \textbf{w/ RGG} & \textbf{Overhead} \\
\midrule
Latency (ms/step) & 1005.6 & 1030.1 & +2.4\% \\
Peak VRAM (GB)    & 32.9 & 35.7 & +8.5\% \\
\bottomrule
\end{tabular}
\end{adjustbox}
\end{center}
\end{table}

\textbf{Convergence Analysis.}
Despite marginal per-step overhead, guided exploration significantly accelerates convergence. As illustrated in \cref{fig:teaser}, {\ours} demonstrates superior sample efficiency by actively steering trajectories toward high-reward regions relative to baselines. Quantitatively, {\ours} attains the peak reward \textbf{1.66}$\times$ faster than DanceGRPO.
This result validates that the gains in exploration efficiency effectively amortize the additional computational cost.


\subsection{Ablation Study}
\label{sec:ablation}

\begin{table}[!tb]
\caption{\textbf{Ablation Study.} We evaluate the impact of active steering by comparing the full {\ours} framework against variants that remove guidance or replace it with filtering strategies.}
\label{tab:ablation}
\begin{center}
\begin{adjustbox}{max width=0.7\linewidth}
\begin{tabular}{l|c}
\toprule
\textbf{Method Setting} & \textbf{VBench2 Total Score} \\
\midrule
\textbf{{\ours} (Full Method)} & \textbf{54.24} \\
\midrule
\quad w/o Active Steering & 53.61 \\
\quad w/o PRM Advantage & 53.95 \\
\quad w/o ORM Advantage & 53.59 \\
\bottomrule
\end{tabular}
\end{adjustbox}
\end{center}
\end{table}

\textbf{Impact of Active Steering.}
As shown in \cref{tab:ablation}, disabling the reward gradient term $\nabla_x r_p$ (w/o Active Steering) causes a performance drop from 54.24 to 53.61. This degradation substantiates the role of actively steering trajectories toward high-reward regions, as undirected stochastic exploration alone proves insufficient for effectively navigating the sparse manifold of high-quality samples.

\textbf{Effectiveness of Dual-Reward Supervision.}
Removing the PRM from advantage computation (w/o PRM Advantage) yields 53.95, indicating that dense step-level feedback enables finer credit assignment beyond the final outcome reward. Removing the ORM instead (w/o ORM Advantage) results in 53.59, a more substantial decline that underscores the importance of pixel-level evaluation for capturing visual fidelity and fine-grained details that the latent-space PRM may overlook. Notably, the ORM ablation exhibits a comparable performance drop to removing active steering entirely, suggesting that outcome-level supervision is critical for grounding the generation process in perceptually meaningful quality metrics. 


\textbf{Sensitivity Analysis of Reward-Gradient Guidance.} We further investigate the hyperparameters governing the RGG mechanism: the guidance coefficient and its temporal activation window.

\textit{1) Guidance Strength.} We investigate the guidance coefficient $\lambda$ governing the RGG mechanism.
We introduce a scalar coefficient $\lambda \geq 0$ to control the magnitude of reward gradient injection:
\begin{equation}
    \mathcal{D}_{\text{total}} = \mathcal{D}_{\text{flow}} + \lambda \nabla_x r_p(X, t).
\end{equation}
$\lambda$ corresponds to $\epsilon_t / \beta$ in \cref{eq:sde_kl_explicit}, 
where $\beta$ is the KL penalty. In practice, we treat $\lambda$ as a tunable hyperparameter 
that balances alignment strength against deviation from the reference flow.
As shown in \cref{tab:rgg_scale}, moderate guidance ($\lambda=0.1$) yields optimal performance. 
Weaker guidance ($\lambda=0.01$) provides insufficient steering toward high-reward regions, 
while stronger guidance ($\lambda=1.0$) causes over-correction that distorts the learned flow dynamics 
and degrades generation quality. 

\begin{table}[!tb]
\caption{\textbf{Ablation on RGG Coefficient ($\lambda$).} We scale the guidance gradient by a factor $\lambda$. Moderate guidance achieves the best trade-off between alignment and quality.}
\label{tab:rgg_scale}
\begin{center}
\begin{adjustbox}{max width=0.60\linewidth}
\begin{tabular}{cc}
\toprule
\textbf{Guidance Scale} ($\lambda$) & \textbf{VBench2 Total} \\
\midrule
0.01 & 53.61 \\
\textbf{0.1} & \textbf{54.24} \\
1.0 & 52.86 \\
\bottomrule
\end{tabular}
\end{adjustbox}
\end{center}
\end{table}

\textit{2) Temporal Activation Window.}
We investigate the temporal sensitivity of RGG across the denoising trajectory.
Although the baseline already achieves a competitive score of 53.61 due to the dual-reward design, applying RGG yields further improvements.
However, full-trajectory guidance provides only marginal gains.
We attribute this to distinct limitations at both ends of the trajectory.
Despite the PRM maintaining consistent accuracy ($>70\%$) across all noise levels shown in \cref{fig:prm_acc}, we observe that applying guidance in the early stage ($t < 0.5$) could disturb the formation of structure and motion.
This aggressive steering induces temporal incoherence, resulting in discontinuous frame transitions that negate the benefits of guidance.
Conversely, while late-stage guidance ($t > 0.75$) avoids these issues, the restricted optimization window provides insufficient steps to accumulate effective corrective updates.
Consequently, the latter half achieves the optimal balance, bypassing the sensitive initialization phase while retaining sufficient capacity for refinement.

\begin{table}[!tb]
\caption{\textbf{Ablation on Temporal Activation.} We compare different guidance windows against the unguided baseline. The results indicate that the latter-half guidance achieves the optimal balance between avoiding early-stage noise and retaining sufficient control authority for refinement.}
\label{tab:rgg_temporal}
\begin{center}
\begin{adjustbox}{max width=0.9\linewidth}
\begin{tabular}{lcc}
\toprule
\textbf{Guidance Window} ($t$) & \textbf{Step Range} & \textbf{VBench2 Total} \\
\midrule
None (Baseline) & - & 53.61 \\
Full Trajectory ($0 \le t \le 1$) & $0 \to 16$ & 53.64 \\
\textbf{Latter Half} ($0.5 \le t \le 1$) & $8 \to 16$ & \textbf{54.24} \\
Late Stage ($0.75 \le t \le 1$) & $12 \to 16$ & 54.14 \\
\bottomrule
\end{tabular}
\end{adjustbox}
\end{center}
\end{table}

\subsection{Analysis of the Process Reward Model}
\label{sec:prm_analysis}

The validity of our methodology relies on the premise that the Latent PRM provides accurate quality assessment even at intermediate noisy states. For reward-gradient guidance to be effective, the PRM must preserve reliable predictions at noisy intermediate states, not merely at clean data.

\begin{figure}[t]
    \centering
    \includegraphics[width=0.8\linewidth]{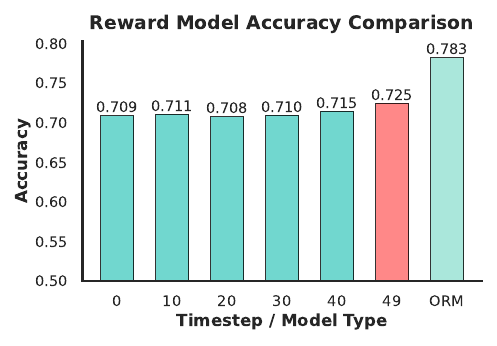}
    \caption{\textbf{Accuracy of the Latent PRM.} We evaluate pairwise accuracy at different timesteps against the pixel-space ORM. The PRM maintains reliable accuracy ($>$70\%), validating its effectiveness for guiding intermediate states.}
    \label{fig:prm_acc}
\end{figure}

As shown in \cref{fig:prm_acc}, our Latent PRM maintains reliable accuracy ($>$70\%) across all evaluated noise levels, with performance approaching that of the pixel-space ORM at lower noise levels. This confirms that the latent-space PRM serves as an effective proxy for visual quality assessment while operating in the latent space.

%% file: section/6-conclusion.tex
\section{Conclusion}
\label{sec:conclusion}

In this paper, we introduced {\ours}, a principled framework that incorporates process reward gradient guided dynamics into video flow matching for efficient post-training alignment.
By deriving a reward-augmented SDE that injects dense gradient signals directly into the flow drift, our method enables active steering toward high-reward regions during exploration, theoretically encompassing prior sampling strategies as special cases.
We further proposed a distillation objective that internalizes the guidance signal into the flow network, eliminating inference-time dependency on the reward model.
Combined with our dual-reward optimization scheme, {\ours} achieves better alignment compared to existing methods while accelerating training convergence by \textbf{1.66}$\times$.

\textbf{Limitations \& Future Work.} 
Our framework relies on a Latent PRM, which presents two constraints.
First, research on latent-space PRMs for video remains in an early stage, and their robustness compared to pixel-based models requires further investigation. 
Second, the PRM is tightly coupled with the specific VAE latent space of the generative backbone, limiting transferability across different video generation architectures.
Future research could explore Representation Autoencoders (RAE)~\cite{zheng2025diffusion}, which utilize fixed, pre-trained visual encoders (e.g., DINOv2) to define a shared latent space.
This offers a promising direction for developing universal, backbone-agnostic PRMs that generalize across video generation models without architecture-specific retraining.

%% file: appendix/appendix_intro.tex
\section*{Appendix}
\begin{itemize}
\item Appendix~\ref{app:proof_score_velocity} presents the formal proof of Proposition~\ref{prop:score_velocity}, establishing the mathematical relationship between the marginal score function and the optimal velocity field under the Linear Optimal Transport path.

\item Appendix~\ref{app:equivalence_flow_grpo} demonstrates that our derived stochastic dynamics naturally recover the sampling procedure proposed in Flow-GRPO, thereby establishing the theoretical foundation underlying these recent methods.

\item Appendix~\ref{app:optimal_policy} provides the complete derivation of the optimal policy that maximizes expected reward subject to a KL divergence constraint, using the method of Lagrange multipliers.

\item Appendix~\ref{app:latent_prm_justification} justifies our design choice of constructing a dedicated Process Reward Model in the latent space, analyzing the computational infeasibility of alternative approaches including decoder backpropagation and zeroth-order gradient estimation.

\item Appendix~\ref{app:logprob_derivation} details the derivation of policy log-probability objectives used in our GRPO formulation, presenting both the standard optimization setting and the policy distillation variant that enables gradient-free inference. This section also includes an empirical comparison of different inference protocols.

\item Appendix~\ref{app:impl_details} provides comprehensive implementation details, including hyperparameter configurations, training infrastructure, and sampling settings used throughout our experiments.

\end{itemize}

%% file: appendix/proof_score_velocity.tex
\section{Proof of Score-Velocity Relationship (Proposition~\ref{prop:score_velocity})}
\label{app:proof_score_velocity}

In this section, we provide the derivation for the relationship between the marginal score function $\nabla \log p_t(x)$ and the optimal vector field $u_t(x)$ under the Linear Optimal Transport (OT) path~\cite{zheng2023guided}.

\textbf{Setup.} Consider the linear interpolation path defined as:
\begin{equation}
    X_t = (1-t)X_0 + tX_1,
\end{equation}
where the source distribution is standard Gaussian, $X_0 \sim p_0 = \mathcal{N}(0, I)$, and $X_1 \sim p_{\text{data}}$.

\textbf{1. Derivation of the Optimal Vector Field $u_t(x)$.}
The optimal vector field $u_t(x)$ that generates the probability path $p_t$ minimizes the flow matching objective. It is known to be the expected instantaneous velocity conditioned on the current location $X_t = x$:
\begin{equation}
    u_t(x) = \mathbb{E}\left[ \frac{d}{dt} X_t \mid X_t = x \right].
\end{equation}
Taking the time derivative of the path yields $\dot{X}_t = X_1 - X_0$. We can express $X_0$ in terms of $X_t$ and $X_1$ as:
\begin{equation}
    X_0 = \frac{X_t - tX_1}{1-t}.
\end{equation}
Substituting this back into the velocity expression:
\begin{equation}
    \dot{X}_t = X_1 - \frac{X_t - tX_1}{1-t} = \frac{(1-t)X_1 - X_t + tX_1}{1-t} = \frac{X_1 - X_t}{1-t}.
\end{equation}
Thus, the conditional vector field is:
\begin{equation} \label{eq:u_expectation}
    u_t(x) = \frac{\mathbb{E}[X_1 \mid X_t = x] - x}{1-t}.
\end{equation}

\textbf{2. Derivation of the Marginal Score $\nabla \log p_t(x)$.}
Conditioned on a fixed data point $X_1 = x_1$, the variable $X_t$ follows a Gaussian distribution because $X_0$ is Gaussian:
\begin{equation}
    p(x_t \mid x_1) = \mathcal{N}(x_t; t x_1, (1-t)^2 I).
\end{equation}
The score of this conditional density is:
\begin{equation}
    \nabla_{x_t} \log p(x_t \mid x_1) = - \frac{x_t - t x_1}{(1-t)^2}.
\end{equation}
Using the identity $\nabla \log p_t(x) = \mathbb{E}_{X_1 \mid X_t=x} [\nabla_x \log p(x \mid X_1)]$, we have:
\begin{equation}
\begin{aligned}
    \nabla \log p_t(x) &= \mathbb{E}\left[ - \frac{x - t X_1}{(1-t)^2} \Bigg| X_t = x \right] \\
    &= - \frac{1}{(1-t)^2} \left( x - t \mathbb{E}[X_1 \mid X_t = x] \right).
\end{aligned}
\end{equation}
Rearranging for the conditional expectation:
\begin{equation} \label{eq:expectation_score}
    \mathbb{E}[X_1 \mid X_t = x] = \frac{1}{t} \left( x + (1-t)^2 \nabla \log p_t(x) \right).
\end{equation}
Note that \cref{eq:expectation_score} is essentially Tweedie's formula adapted for the linear OT path.

\textbf{3. Connecting Score and Velocity.}
We substitute the term $\mathbb{E}[X_1 \mid X_t = x]$ from \cref{eq:u_expectation} into the score equation. From \cref{eq:u_expectation}, we have:
\begin{equation}
    \mathbb{E}[X_1 \mid X_t = x] = x + (1-t) u_t(x).
\end{equation}
Substitute this into the score expression:
\begin{equation}
\begin{aligned}
    \nabla \log p_t(x) &= - \frac{1}{(1-t)^2} \left( x - t \left[ x + (1-t) u_t(x) \right] \right) \\
    &= - \frac{1}{(1-t)^2} \left( x - tx - t(1-t) u_t(x) \right) \\
    &= - \frac{1}{(1-t)^2} \left( (1-t)x - t(1-t) u_t(x) \right) \\
    &= - \frac{1}{1-t} \left( x - t u_t(x) \right).
\end{aligned}
\end{equation}
\qed

%% file: appendix/equivalence_flow_grpo.tex
\section{Equivalence to Flow-GRPO Dynamics}
\label{app:equivalence_flow_grpo}
In this appendix, we demonstrate that the specific stochastic dynamics derived in \cref{eq:sde_density_explicit} naturally recover the dynamics proposed in recent works such as Flow-GRPO. By showing this mathematical alignment, we establish that our framework provides the underlying mathematical foundation for these sampling methods.

\paragraph{Notation Mapping.}
We distinguish between the variables used in our framework and those in Flow-GRPO as follows:
\begin{itemize}
    \item \textbf{Time Convention:} 
    Flow-GRPO defines time $\tau \in [0, 1]$ (reverse process, $\tau=0$ is data). In our settings, we define time $t \in [0, 1]$ (forward process, $t=1$ is data). The relationship is:
    \begin{equation}
        t = 1 - \tau, \quad dt = -d\tau.
    \end{equation}
    
    \item \textbf{State Variable:} 
    Let $x_\tau$ denote the state in Flow-GRPO at time $\tau$, and $X_t$ denote the state in our framework at time $t$. Since they represent the same physical trajectory viewed from different temporal directions, the correspondence is:
    \begin{equation}
        X_{1-\tau} \equiv x_\tau.
    \end{equation}
    
    \item \textbf{Velocity Field:} 
    Let $v_\tau(x_\tau)$ be the velocity in Flow-GRPO (pointing data $\to$ noise). Our velocity $u_t(X_t)$ points noise $\to$ data. At the corresponding physical state, these vectors are opposite:
    \begin{equation}
        u_{1-\tau}(X_{1-\tau}) = -v_\tau(x_\tau).
    \end{equation}
    
    \item \textbf{Noise Level:} 
    Flow-GRPO uses diffusion coefficient $\sigma_\tau$. Mapping this to our noise schedule $\epsilon_t$ at the equivalent time $t=1-\tau$:
    \begin{equation}
        \epsilon_{1-\tau} = \frac{1}{2}\sigma_\tau^2.
    \end{equation}
\end{itemize}

\paragraph{Derivation.}
The Flow-GRPO SDE for generation is given by:
\begin{equation}
    dx_\tau = \left[ v_\tau(x_\tau) + \frac{\sigma_\tau^2}{2\tau}\left( x_\tau + (1-\tau)v_\tau(x_\tau) \right) \right] d\tau + \sigma_\tau dw.
\end{equation}
To prove equivalence, we start with our derived drift term $\mathcal{D}_t$ for the state $X_t$:
\begin{equation}
    \mathcal{D}_t = \left(1 + \frac{t \epsilon_t}{1-t}\right)u_t(X_t) - \frac{\epsilon_t}{1-t}X_t.
\end{equation}
We proceed step-by-step. First, we substitute the time variable $t = 1-\tau$ into our expression, while keeping the function forms generic:
\begin{align}
    \mathcal{D}_{1-\tau} &= \left(1 + \frac{(1-\tau) \epsilon_{1-\tau}}{1-(1-\tau)}\right)u_{1-\tau}(X_{1-\tau}) - \frac{\epsilon_{1-\tau}}{1-(1-\tau)}X_{1-\tau} \nonumber \\
    &= \left(1 + \frac{(1-\tau) \epsilon_{1-\tau}}{\tau}\right)u_{1-\tau}(X_{1-\tau}) - \frac{\epsilon_{1-\tau}}{\tau}X_{1-\tau}.
\end{align}
Next, we apply the notation mappings: $X_{1-\tau} \to x_\tau$, $u_{1-\tau}(X_{1-\tau}) \to -v_\tau(x_\tau)$, and $\epsilon_{1-\tau} \to \sigma_\tau^2/2$:
\begin{align}
    \mathcal{D}_{1-\tau} &= \left(1 + \frac{(1-\tau) (\sigma_\tau^2/2)}{\tau}\right)\left( -v_\tau(x_\tau) \right) - \frac{\sigma_\tau^2/2}{\tau}x_\tau \nonumber \\
    &= -v_\tau(x_\tau) - \frac{(1-\tau)\sigma_\tau^2}{2\tau}v_\tau(x_\tau) - \frac{\sigma_\tau^2}{2\tau}x_\tau \nonumber \\
    &= - \left[ v_\tau(x_\tau) + \frac{\sigma_\tau^2}{2\tau}\left( x_\tau + (1-\tau)v_\tau(x_\tau) \right) \right].
\end{align}
Finally, considering the differential time relation $dt = -d\tau$, the infinitesimal update in our framework becomes:
\begin{equation}
    dX_t = \mathcal{D}_t dt = \mathcal{D}_{1-\tau} (-d\tau) = \left[ v_\tau(x_\tau) + \frac{\sigma_\tau^2}{2\tau}\left( x_\tau + (1-\tau)v_\tau(x_\tau) \right) \right] d\tau.
\end{equation}
This matches the drift term of the Flow-GRPO SDE exactly, confirming that both frameworks describe the same underlying stochastic process.

%% file: appendix/optimal_policy.tex
\section{Derivation of the Optimal Policy}
\label{app:optimal_policy}

In this section, we provide the formal derivation for the optimal policy $\pi^*$ that maximizes the expected reward subject to a KL divergence constraint. Following the framework of Direct Preference Optimization (DPO) \cite{rafailov2023direct}, we employ the method of Lagrange multipliers to solve for the optimal distribution directly without assuming a candidate form.

\textbf{Problem Statement.} 
We aim to determine a policy $\pi(x)$ that maximizes the following objective functional:
\begin{equation}
    \mathcal{J}(\pi) = \mathbb{E}_{x \sim \pi}[r(x)] - \beta D_{\text{KL}}(\pi \| \pi_{\text{ref}}) = \int \pi(x) r(x) dx - \beta \int \pi(x) \log \frac{\pi(x)}{\pi_{\text{ref}}(x)} dx,
\end{equation}
subject to the normalization constraint ensuring $\pi(x)$ is a valid probability density function:
\begin{equation}
    \mathcal{C}(\pi) = \int \pi(x) dx - 1 = 0.
\end{equation}

\textbf{Derivation.}
We construct the Lagrangian functional $\mathcal{L}(\pi, \lambda)$, where $\lambda$ is the Lagrange multiplier associated with the normalization constraint:
\begin{equation}
    \mathcal{L}(\pi, \lambda) = \mathcal{J}(\pi) - \lambda \left( \int \pi(x) dx - 1 \right).
\end{equation}
Expanding the terms, we have:
\begin{equation}
    \mathcal{L}(\pi, \lambda) = \int \pi(x) \left( r(x) - \beta \log \pi(x) + \beta \log \pi_{\text{ref}}(x) - \lambda \right) dx + \lambda.
\end{equation}

To find the stationary point $\pi^*$, we take the functional derivative of $\mathcal{L}$ with respect to $\pi(x)$ and set it to zero (the Euler-Lagrange condition):
\begin{equation}
    \frac{\delta \mathcal{L}}{\delta \pi(x)} = r(x) - \beta (1 + \log \pi(x)) + \beta \log \pi_{\text{ref}}(x) - \lambda = 0.
\end{equation}
Note that $\frac{\delta}{\delta \pi}(\pi \log \pi) = 1 + \log \pi$. Rearranging the terms to solve for $\log \pi(x)$:
\begin{equation}
    \beta \log \pi(x) = r(x) + \beta \log \pi_{\text{ref}}(x) - (\beta + \lambda).
\end{equation}
Dividing by $\beta$ and exponentiating both sides, we obtain the general form of the optimal policy:
\begin{equation}
    \pi^*(x) = \pi_{\text{ref}}(x) \exp\left( \frac{r(x)}{\beta} \right) \exp\left( -1 - \frac{\lambda}{\beta} \right).
\end{equation}

The term $\exp\left( -1 - \frac{\lambda}{\beta} \right)$ is independent of $x$ and serves as the normalization constant. Let $Z^{-1} = \exp\left( -1 - \frac{\lambda}{\beta} \right)$. Using the constraint $\int \pi^*(x) dx = 1$, we determine $Z$:
\begin{equation}
    \int \frac{1}{Z} \pi_{\text{ref}}(x) \exp\left( \frac{r(x)}{\beta} \right) dx = 1 \implies Z = \int \pi_{\text{ref}}(x) \exp\left( \frac{r(x)}{\beta} \right) dx.
\end{equation}

Thus, the optimal policy is uniquely determined as:
\begin{equation}
    \pi^*(x) = \frac{1}{Z} \pi_{\text{ref}}(x) \exp\left( \frac{r(x)}{\beta} \right).
\end{equation}
\qed

%% file: appendix/latent_prm_justification.tex
\section{Justification for Latent-Space Guidance}
\label{app:latent_prm_justification}

In \cref{sec:prm_training}, we construct a dedicated PRM operating in the latent space. While utilizing off-the-shelf pixel-space reward models (e.g., VideoAlign~\cite{liu2025improving}) is theoretically appealing, it presents significant computational challenges for online optimization. We analyze two alternative integration strategies and detail why both are impractical for our framework: direct backpropagation via the decoder and zeroth-order gradient estimation.

\subsection{Memory Constraints of Decoder Backpropagation}

A direct approach to guide the latent $x_t$ is to chain the generation and scoring processes. Based on the Linear Optimal Transport path defined in \cref{sec:prelim}, the clean latent $\hat{x}_1$ can be estimated from the current noisy state $x_t$ via the vector field $u_\theta$:
\begin{equation}
    \hat{x}_1 = x_t + (1-t)u_\theta(x_t, t).
\end{equation}
Let $\hat{v} = \mathcal{D}(\hat{x}_1)$ denote the decoded video pixels. The gradient of the reward $s = r_{\text{pixel}}(\hat{v})$ with respect to $x_t$ is given by the chain rule:
\begin{equation} \label{eq:chain_rule}
    \nabla_{x_t} s = \underbrace{\left(\frac{\partial \hat{x}_1}{\partial x_t}\right)^\top}_{\text{Denoise Grad}} \cdot \underbrace{\left(\frac{\partial \mathcal{D}}{\partial \hat{x}_1}\right)^\top}_{\text{Decoder Jacobian}} \cdot \underbrace{\nabla_{\hat{v}} r_{\text{pixel}}}_{\text{Reward Grad}}
\end{equation}
The bottleneck lies in the Decoder Jacobian term. Video VAEs employ high spatial-temporal compression ratios (e.g., $8 \times 8 \times 4$), resulting in pixel-space tensors $\hat{v}$ that are orders of magnitude larger than the latent $x_t$. Backpropagating through the decoder requires storing dense activation maps for the entire video sequence, which exceeds the VRAM capacity of typical training GPUs and causes Out-Of-Memory errors.

\subsection{Variance Issues in Zeroth-Order Estimation}

To circumvent memory constraints, one might employ zeroth-order (derivative-free) optimization such as Simultaneous Perturbation Stochastic Approximation (SPSA), which estimates gradients using only forward passes without storing activation graphs.

For a composite reward function $F(x_t) = r_{\text{pixel}}(\mathcal{D}(\text{denoise}(x_t)))$, the gradient can be approximated via two-sided perturbation with a random vector $z \sim \mathcal{N}(0, I)$:
\begin{equation} \label{eq:spsa_est}
    \hat{g}_{\text{SPSA}}(x_t) \approx \frac{F(x_t + \sigma z) - F(x_t - \sigma z)}{2\sigma} z,
\end{equation}
where $\sigma$ is the perturbation scale. While memory-efficient, this estimator suffers from high variance in high-dimensional latent spaces. Obtaining a reliable descent direction typically requires averaging over $K$ independent perturbations, where larger $K$ reduces variance but proportionally increases computational cost.

In our experiments, we attempted gradient estimation with minimal sampling ($K=1$), using a single sample perturbed twice to obtain positive and negative variants. The resulting videos exhibited severe degradation, with visual content collapsing into block-shaped noise patterns. This failure mode indicates that single-sample SPSA provides insufficient gradient signal for meaningful optimization in high-dimensional video latent spaces. Increasing $K$ to achieve acceptable variance (e.g., $K \ge 10$) would require decoding and scoring dozens of video candidates at every timestep, introducing prohibitive latency.

\subsection{Summary}

Given that backpropagation is strictly memory-bound and accurate zeroth-order estimation is time-bound, training a dedicated PRM directly in the latent space offers the necessary balance between gradient quality and computational efficiency.

%% file: appendix/logprob_derivation.tex
\section{Derivation of Policy Log-Probability Objectives}
\label{app:logprob_derivation}

In this section, we provide the detailed derivation of the log-probability $\log \pi_\theta(X^i_{k+1} | X^i_k)$ used to compute the importance sampling ratio $\omega_{k, i}(\theta)$ in the GRPO objective. We present two formulations corresponding to different inference requirements after training: (1) Standard Optimization, where reward gradients are explicitly used during both training and inference; and (2) Policy Distillation, where the reward guidance is distilled into the flow network.

\subsection{Standard Gradient-Guided Optimization}
\label{app:standard_deriv}

We first derive the objective for the standard setting. In this setting, both the behavior policy used for sampling and the target policy being optimized are defined to explicitly incorporate the Reward-Gradient Guidance (RGG).

Let $\psi$ denote the policy parameters (where $\psi \in \{\theta, \theta_{\text{old}}\}$). The transition mean $\mu_\psi(X_k, t_k)$ from state $X_k$ to $X_{k+1}$ is defined as the current state plus the total drift:
\begin{equation} \label{eq:standard_mean_def}
    \mu_\psi(X_k, t_k) = X_k + \Delta t \cdot \mathcal{D}_{\text{total}}(X_k, t_k; \psi).
\end{equation}
The total drift $\mathcal{D}_{\text{total}}$ consists of the learnable flow field and the fixed reward guidance:
\begin{equation}
    \mathcal{D}_{\text{total}}(X, t; \psi) = \underbrace{\left[\left(1 + \frac{t \epsilon_t}{1-t}\right)u_\psi(X, t) - \frac{\epsilon_t}{1-t}X\right]}_{\text{Flow Drift } \mathcal{D}_{\text{flow}}} + \underbrace{\frac{\epsilon_t}{\beta} \nabla_x r_p(X, t)}_{\text{Reward Guidance}}.
\end{equation}

\noindent\textbf{Phase 1: Sampling (Data Collection).}
During exploration, we generate trajectories using the frozen behavior policy $\theta_{\text{old}}$. The next state $X_{k+1}$ is sampled via:
\begin{equation} \label{eq:sampling_step_standard}
    X_{k+1} = \mu_{\theta_{\text{old}}}(X_k, t_k) + \sqrt{2\epsilon_{t_k} \Delta t} Z_k,
\end{equation}
where $Z_k \sim \mathcal{N}(0, I)$ is standard Gaussian noise.
The log-probability of generating this specific sample is ($C_k$ is a timestep-dependent constant):
\begin{equation}
\begin{aligned}
    \log \pi_{\theta_{\text{old}}}(X_{k+1} | X_k) &= C_k - \frac{1}{4\epsilon_{t_k} \Delta t} \left\| X_{k+1} - \mu_{\theta_{\text{old}}}(X_k, t_k) \right\|^2 \\
    &= C_k - \frac{1}{4\epsilon_{t_k} \Delta t} \left\| \sqrt{2\epsilon_{t_k} \Delta t} Z_k \right\|^2 \\
    &= C_k - \frac{1}{2} \| Z_k \|^2.
\end{aligned}
\end{equation}

\noindent\textbf{Phase 2: Optimization Objective.}
During the GRPO update, we maximize the log-probability of the target policy $\pi_\theta$ generating the sample $X_{k+1}$. The log-probability is given by:
\begin{equation} \label{eq:target_logp_standard}
    \log \pi_\theta(X_{k+1} | X_k) = C_k - \frac{1}{4\epsilon_{t_k} \Delta t} \left\| X_{k+1} - \mu_\theta(X_k, t_k) \right\|^2.
\end{equation}
Substituting \cref{eq:sampling_step_standard} into the residual $(X_{k+1} - \mu_\theta)$ and expanding the drift definitions reveals that the reward guidance terms cancel out:
\begin{equation}
\begin{aligned}
    X_{k+1} - \mu_\theta(X_k, t_k) &= \left( \mu_{\theta_{\text{old}}}(X_k, t_k) + \sqrt{2\epsilon_{t_k} \Delta t} Z_k \right) - \mu_\theta(X_k, t_k) \\
    &= \sqrt{2\epsilon_{t_k} \Delta t} Z_k + \left( \mu_{\theta_{\text{old}}}(X_k, t_k) - \mu_\theta(X_k, t_k) \right) \\
    &= \sqrt{2\epsilon_{t_k} \Delta t} Z_k + \Delta t \left( \mathcal{D}_{\text{total}}(X_k; \theta_{\text{old}}) - \mathcal{D}_{\text{total}}(X_k; \theta) \right) \\
    &= \sqrt{2\epsilon_{t_k} \Delta t} Z_k + \Delta t \left( \left[\mathcal{D}_{\text{flow}}^{\text{old}} + \frac{\epsilon}{\beta}\nabla r_p\right] - \left[\mathcal{D}_{\text{flow}}^{\theta} + \frac{\epsilon}{\beta}\nabla r_p\right] \right) \\
    &= \sqrt{2\epsilon_{t_k} \Delta t} Z_k + \Delta t \left( \left[ \left(1 + \frac{t\epsilon}{1-t}\right)u_{\theta_{\text{old}}} - \frac{\epsilon}{1-t}X_k \right] - \left[ \left(1 + \frac{t\epsilon}{1-t}\right)u_{\theta} - \frac{\epsilon}{1-t}X_k \right] \right) \\
    &= \sqrt{2\epsilon_{t_k} \Delta t} Z_k + \Delta t \left(1 + \frac{t_k \epsilon_{t_k}}{1-t_k}\right) \left( u_{\theta_{\text{old}}}(X_k, t_k) - u_\theta(X_k, t_k) \right).
\end{aligned}
\end{equation}

\textbf{Remark on Inference/Deployment.}
From a theoretical standpoint, the SDE formulation implies that the reward gradient $\nabla_x r_p$ constitutes an intrinsic component of the drift, suggesting it should be retained during inference to strictly match the training dynamics.
However, retaining this term for deployment introduces a significant practical bottleneck: it necessitates the concurrent deployment of the auxiliary PRM alongside the video generator, increasing system complexity and memory footprint.
Empirically, we observe that through the GRPO optimization process, the flow network $u_\theta$ implicitly learns to approximate the guided field.
Consequently, for standard deployment (which we term the Implicit Mode), we can discard the external reward gradient and rely solely on the learned weights of $u_\theta$.
This strategy eliminates the dependency on the external LRM while maintaining high alignment performance.

\subsection{Policy Distillation for Gradient-Free Inference}
\label{app:distillation_formulation}

We now consider the Policy Distillation scenario. In this formulation, we treat the reward-guided trajectory generation as a ``teacher" process and optimize the ``student" flow network to internalize this guidance, enabling gradient-free inference.

\noindent\textbf{Teacher Process (Data Collection \& Distribution Shift).}
First, we analyze the behavior of the teacher model. We generate trajectories using the explicitly guided behavior policy, but we evaluate the sample's probability under the \textit{unguided} behavior policy to quantify the distribution shift induced by the reward.

The unguided teacher mean $\mu_{\theta_{\text{old}}}$ is defined purely by the base flow dynamics:
\begin{equation}
    \mu_{\theta_{\text{old}}}(X_k, t_k) = X_k + \Delta t \cdot \mathcal{D}_{\text{flow}}(X_k, t_k; \theta_{\text{old}}).
\end{equation}
However, the sample $X_{k+1}$ is generated using the Guided Dynamics:
\begin{equation} \label{eq:distill_sample_gen}
    X_{k+1} = X_k + \Delta t \left[ \mathcal{D}_{\text{flow}}(X_k, t_k; \theta_{\text{old}}) + \frac{\epsilon_{t_k}}{\beta} \nabla_x r_p(X_k, t_k) \right] + \sqrt{2\epsilon_{t_k} \Delta t} Z_k.
\end{equation}
We derive the log-probability of this guided sample under the unguided teacher policy:
\begin{equation}
    \log \pi_{\theta_{\text{old}}}(X_{k+1} | X_k) = C_k - \frac{1}{4\epsilon_{t_k} \Delta t} \left\| X_{k+1} - \mu_{\theta_{\text{old}}}(X_k, t_k) \right\|^2.
\end{equation}
Substituting \cref{eq:distill_sample_gen} into the residual term, we observe the explicit shift:
\begin{equation}
\begin{aligned}
    X_{k+1} - \mu_{\theta_{\text{old}}}(X_k, t_k) &= \left( X_k + \Delta t \left[ \mathcal{D}_{\text{flow}}^{\text{old}} + \frac{\epsilon_{t_k}}{\beta} \nabla_x r_p \right] + \sqrt{2\epsilon_{t_k} \Delta t} Z_k \right) - \left( X_k + \Delta t \mathcal{D}_{\text{flow}}^{\text{old}} \right) \\
    &= \sqrt{2\epsilon_{t_k} \Delta t} Z_k + \Delta t \left( \mathcal{D}_{\text{flow}}(X_k, t_k; \theta_{\text{old}}) - \mathcal{D}_{\text{flow}}(X_k, t_k; \theta_{\text{old}}) + \frac{\epsilon_{t_k}}{\beta} \nabla_x r_p(X_k, t_k) \right) \\
    &= \sqrt{2\epsilon_{t_k} \Delta t} Z_k + \Delta t \left( \frac{\epsilon_{t_k}}{\beta} \nabla_x r_p(X_k, t_k) \right).
\end{aligned}
\end{equation}
This derivation mathematically confirms that the guided sample $X_{k+1}$ deviates from the unguided teacher's expectation exactly by the reward gradient term.

\noindent\textbf{Student Process (Optimization Objective).} The student policy is defined \textbf{without} explicit reward guidance:
\begin{equation}
    \mu_\theta(X_k, t_k) = X_k + \Delta t \cdot \mathcal{D}_{\text{flow}}(X_k, t_k; \theta).
\end{equation}
\begin{equation} \label{eq:logp_distill}
    \log \pi_\theta(X_{k+1} | X_k) = C_k - \frac{1}{4\epsilon_{t_k} \Delta t} \left\| X_{k+1} - \mu_\theta(X_k, t_k) \right\|^2.
\end{equation}
We substitute the same sample $X_{k+1}$ from \cref{eq:distill_sample_gen} into this residual term:
\begin{equation}
\begin{aligned}
    X_{k+1} - \mu_\theta(X_k, t_k) &= \left( X_k + \Delta t \left[ \mathcal{D}_{\text{flow}}^{\text{old}} + \frac{\epsilon_{t_k}}{\beta} \nabla_x r_p \right] + \sqrt{2\epsilon_{t_k} \Delta t} Z_k \right) - \left( X_k + \Delta t \mathcal{D}_{\text{flow}}^{\theta} \right) \\
    &= \sqrt{2\epsilon_{t_k} \Delta t} Z_k + \Delta t \left( \mathcal{D}_{\text{flow}}(X_k, t_k; \theta_{\text{old}}) - \mathcal{D}_{\text{flow}}(X_k, t_k; \theta) + \frac{\epsilon_{t_k}}{\beta} \nabla_x r_p(X_k, t_k) \right) \\
    &= \sqrt{2\epsilon_{t_k} \Delta t} Z_k + \Delta t \left( \left[ \left(1 + \frac{t_k \epsilon_{t_k}}{1-t_k}\right)u_{\theta_{\text{old}}} - \frac{\epsilon_{t_k}}{1-t_k}X_k \right] - \left[ \left(1 + \frac{t_k \epsilon_{t_k}}{1-t_k}\right)u_{\theta} - \frac{\epsilon_{t_k}}{1-t_k}X_k \right] + \frac{\epsilon_{t_k}}{\beta} \nabla_x r_p \right) \\
    &= \sqrt{2\epsilon_{t_k} \Delta t} Z_k + \Delta t \left[ \left(1 + \frac{t_k \epsilon_{t_k}}{1-t_k}\right) \left( u_{\theta_{\text{old}}}(X_k, t_k) - u_\theta(X_k, t_k) \right) + \frac{\epsilon_{t_k}}{\beta} \nabla_x r_p(X_k, t_k) \right].
\end{aligned}
\end{equation}
The equation shows that the student velocity $u_\theta$ is forced to match the behavior velocity $u_{\theta_{\text{old}}}$ \textit{plus} the direction induced by the process reward gradient $\nabla_x r_p$. This effectively distills the reward information directly into the weights of the flow network.

\textbf{Remark on Inference/Deployment.}
In this distillation framework, the optimization objective explicitly forces the student flow network $u_\theta$ to internalize the guidance signal provided by the teacher.
Consequently, the student model functions independently in deployment.
It generates high-reward trajectories relying solely on its learned velocity field, thereby \textit{completely eliminating} the need to compute $\nabla_x r_p$ or load the reward model.
This results in a streamlined inference process that is operationally identical to the base model but with aligned behavior.

\subsection{Empirical Comparison of Inference Protocols}
\label{app:logprob_empirical}

We evaluate three distinct inference protocols derived from our framework to understand the trade-offs between alignment quality, computational cost, and deployment complexity.
All variants utilize the same HunyuanVideo-14B backbone.

\begin{enumerate}
    \item \textbf{Inference RGG:} Using the Standard Optimization model and \textit{retaining} the RGG ($\nabla_x r_p$) during inference. This requires loading the LRM.
    \item \textbf{Implicit:} Using the Standard Optimization model but \textit{discarding} the RGG during inference.
    \item \textbf{Distilled:} Using the model trained via the Policy Distillation objective, sampled without RGG.
\end{enumerate}

\begin{table}[h]
\caption{\textbf{Comparison of Inference Protocols.} 
We compare three variants derived from our framework. 
Ours (Distilled) achieves the best alignment performance by explicitly supervising the flow network to mimic the guided trajectory. 
Crucially, it eliminates the need for an external LRM during inference, offering the best trade-off between quality and efficiency. Note that the Inference RGG variant encounters out-of-memory (OOM) errors on a single H20 GPU at high resolutions and frame counts ($81 \times 640 \times 640$), rendering it impractical for standard deployment scenarios.}
\label{tab:logp_comparison}
\begin{center}
\begin{adjustbox}{max width=\linewidth}
\begin{tabular}{lccccc}
\toprule
\textbf{Method Variant} & \textbf{Training Objective} & \textbf{Inference Guidance} & \textbf{External LRM} & \textbf{VBench2 Total} \\
\midrule
Ours (Implicit) & Standard (\cref{eq:target_logp_standard}) & w/o RGG & None  & 53.71 \\
Ours (Inference RGG) & Standard (\cref{eq:target_logp_standard}) & w/ RGG ($\lambda=0.1$) & \textbf{Required}  & OOM \\
\textbf{Ours (Distilled)} & Distill (\cref{eq:logp_distill}) & \textbf{w/o RGG} & \textbf{None}  & \textbf{54.24} \\
\bottomrule
\end{tabular}
\end{adjustbox}
\end{center}
\end{table}

\textbf{Analysis.} 
As presented in \cref{tab:logp_comparison}, the \textbf{Policy Distillation} variant demonstrates better performance than the standard implicit baseline. 
Notably, the ``Inference RGG'' mode, which retains the RGG during inference, encounters out-of-memory (OOM) errors when deployed on a single NVIDIA H20 GPU (96 GB VRAM) without model parallelism. 
This limitation arises from the substantial memory overhead introduced by concurrently loading both the video generation backbone and the external LRM, compounded by the need to compute and backpropagate reward gradients through high-resolution, high-frame-count video latents.
This practical constraint underscores the critical importance of the distillation approach: the distilled model explicitly internalizes the preference signal into its velocity field via the student-teacher objective, enabling deployment with the same memory footprint as the base model.
This explicit distillation further enables the model to capture the guided dynamics more accurately than the ``Implicit'' baseline, achieving state-of-the-art alignment scores. 
Given that it eliminates the dependency on the external LRM, avoids the prohibitive memory requirements of runtime guidance, and delivers the highest generation quality, we adopt the \textbf{Distilled} formulation as our primary method for all experiments.

%% file: appendix/implementation_details.tex
\section{Implementation Details.}
\label{app:impl_details}
Our framework is implemented using PyTorch~\cite{paszke2019pytorch}. 
For distributed training, we utilize Fully Sharded Data Parallel (FSDP)~\cite{zhao2023pytorch} with a "Full Sharding" strategy to efficiently manage the memory footprint of the 14B-parameter backbone. 
The training is conducted on a high-performance cluster consisting of 5 nodes, each equipped with 8 NVIDIA H800 GPUs (80GB VRAM) and an Intel(R) Xeon(R) Platinum 8476C CPU. 
We employ gradient checkpointing and mixed-precision training (bfloat16) to further optimize resource usage. 
With a per-device batch size of 1 and gradient accumulation steps of 4 across 40 GPUs, the effective global batch size is 160.

\begin{table*}[h]
\caption{\textbf{Hyperparameters and Implementation Details.} We list the detailed configuration used for training {\ours} with the HunyuanVideo-14B backbone. Note that the sampling configuration differs between the training rollout phase and the final evaluation.}
\label{tab:impl_details}
\begin{center}
\begin{adjustbox}{max width=0.95\textwidth}
\begin{tabular}{l|l|c|c}
\toprule
\textbf{Category} & \textbf{Hyperparameter} & \textbf{Symbol} & \textbf{Value} \\
\midrule
\multirow{4}{*}{\textbf{Model Architecture}} 
& Backbone Model & $u_\theta$ & HunyuanVideo~\cite{kong2024hunyuanvideo}\\
& Process Reward Model (PRM) & $r_p$ & Latent DiT (8 Layers) \\
& Outcome Reward Model (ORM) & $r_o$ & InternVL3-1B~\cite{zhu2025internvl3} \\
& VAE Compression Factor & - & $8 \times 8 \times 4$ \\
\midrule
\multirow{8}{*}{\textbf{Training Configuration}} 
& Optimizer & - & AdamW~\cite{loshchilov2017decoupled} \\
& Precision & - & bfloat16 \\
& Learning Rate & $\eta$ & $1 \times 10^{-6}$ \\
& Weight Decay & - & $1 \times 10^{-4}$ \\
& Batch Size (per Device) & - & 1 \\
& Gradient Accumulation Steps & - & 4 \\
& Total GPUs & - & 40 \\
& Total Training Steps & - & 250 \\
\midrule
\multirow{5}{*}{\textbf{Sampling (Training Rollout)}} 
& Sampling Steps & $T$ & 16 \\
& Group Size (per prompt) & $G$ & 8 \\
& Resolution & $H \times W$ & $480 \times 480$ \\
& Number of Frames & $F$ & 32 \\
& Time Shift & $s$ & 5.0 \\
\midrule
\multirow{3}{*}{\textbf{Sampling (Evaluation)}} 
& Sampling Steps & $T$ & 30 \\
& Resolution & $H \times W$ & $640 \times 640$ \\
& Number of Frames & $F$ & 81 \\
\midrule
\multirow{5}{*}{\textbf{Reward-Gradient Guidance}} 
& Guidance Scale & $\lambda$ & 0.1 \\
& KL Regularization Coefficient & $\beta$ & 3.125 \\
& Exploration Noise & $\epsilon_t$ & 0.3125 \\
& Guidance Window (Steps) & - & $8 \to 16$ (Latter Half) \\
& Guidance Window (Time) & $t$ & $[0.5, 1.0]$ \\
\midrule
\multirow{3}{*}{\textbf{GRPO Optimization}} 
& Process Reward Weight & - & 1.0 \\
& Outcome Reward Weight & - & 1.0 \\
& Clip Parameter & $\varepsilon_{\text{clip}}$ & $1 \times 10^{-4}$ \\
\bottomrule
\end{tabular}
\end{adjustbox}
\end{center}
\end{table*}

%% file: ref.bib
@article{lipman2022flow,
  title={Flow matching for generative modeling},
  author={Lipman, Yaron and Chen, Ricky TQ and Ben-Hamu, Heli and Nickel, Maximilian and Le, Matt},
  journal={arXiv preprint arXiv:2210.02747},
  year={2022}
}

@article{tong2023improving,
  title={Improving and generalizing flow-based generative models with minibatch optimal transport},
  author={Tong, Alexander and Fatras, Kilian and Malkin, Nikolay and Huguet, Guillaume and Zhang, Yanlei and Rector-Brooks, Jarrid and Wolf, Guy and Bengio, Yoshua},
  journal={arXiv preprint arXiv:2302.00482},
  year={2023}
}

@article{liu2022flow,
  title={Flow straight and fast: Learning to generate and transfer data with rectified flow},
  author={Liu, Xingchao and Gong, Chengyue and Liu, Qiang},
  journal={arXiv preprint arXiv:2209.03003},
  year={2022}
}

@article{bradley1952rank,
  title={Rank analysis of incomplete block designs: I. the method of paired comparisons},
  author={Bradley, Ralph Allan and Terry, Milton E},
  journal={Biometrika},
  volume={39},
  number={3/4},
  pages={324--345},
  year={1952},
  publisher={JSTOR}
}

@article{zhu2025internvl3,
  title={Internvl3: Exploring advanced training and test-time recipes for open-source multimodal models},
  author={Zhu, Jinguo and Wang, Weiyun and Chen, Zhe and Liu, Zhaoyang and Ye, Shenglong and Gu, Lixin and Tian, Hao and Duan, Yuchen and Su, Weijie and Shao, Jie and others},
  journal={arXiv preprint arXiv:2504.10479},
  year={2025}
}

@article{albergo2023stochastic,
  title={Stochastic interpolants: A unifying framework for flows and diffusions},
  author={Albergo, Michael S and Boffi, Nicholas M and Vanden-Eijnden, Eric},
  journal={arXiv preprint arXiv:2303.08797},
  year={2023}
}

@article{albergo2024nets,
  title={Nets: A non-equilibrium transport sampler},
  author={Albergo, Michael S and Vanden-Eijnden, Eric},
  journal={arXiv preprint arXiv:2410.02711},
  year={2024}
}

@article{kong2024hunyuanvideo,
  title={Hunyuanvideo: A systematic framework for large video generative models},
  author={Kong, Weijie and Tian, Qi and Zhang, Zijian and Min, Rox and Dai, Zuozhuo and Zhou, Jin and Xiong, Jiangfeng and Li, Xin and Wu, Bo and Zhang, Jianwei and others},
  journal={arXiv preprint arXiv:2412.03603},
  year={2024}
}

@inproceedings{esser2024scaling,
  title={Scaling rectified flow transformers for high-resolution image synthesis},
  author={Esser, Patrick and Kulal, Sumith and Blattmann, Andreas and Entezari, Rahim and M{\"u}ller, Jonas and Saini, Harry and Levi, Yam and Lorenz, Dominik and Sauer, Axel and Boesel, Frederic and others},
  booktitle={Forty-first international conference on machine learning},
  year={2024}
}

@article{black2023training,
  title={Training diffusion models with reinforcement learning},
  author={Black, Kevin and Janner, Michael and Du, Yilun and Kostrikov, Ilya and Levine, Sergey},
  journal={arXiv preprint arXiv:2305.13301},
  year={2023}
}

@article{liu2025flow,
  title={Flow-grpo: Training flow matching models via online rl},
  author={Liu, Jie and Liu, Gongye and Liang, Jiajun and Li, Yangguang and Liu, Jiaheng and Wang, Xintao and Wan, Pengfei and Zhang, Di and Ouyang, Wanli},
  journal={arXiv preprint arXiv:2505.05470},
  year={2025}
}

@article{xue2025dancegrpo,
  title={DanceGRPO: Unleashing GRPO on Visual Generation},
  author={Xue, Zeyue and Wu, Jie and Gao, Yu and Kong, Fangyuan and Zhu, Lingting and Chen, Mengzhao and Liu, Zhiheng and Liu, Wei and Guo, Qiushan and Huang, Weilin and others},
  journal={arXiv preprint arXiv:2505.07818},
  year={2025}
}

@article{ho2022classifier,
  title={Classifier-free diffusion guidance},
  author={Ho, Jonathan and Salimans, Tim},
  journal={arXiv preprint arXiv:2207.12598},
  year={2022}
}

@inproceedings{lu2023contrastive,
  title={Contrastive energy prediction for exact energy-guided diffusion sampling in offline reinforcement learning},
  author={Lu, Cheng and Chen, Huayu and Chen, Jianfei and Su, Hang and Li, Chongxuan and Zhu, Jun},
  booktitle={International Conference on Machine Learning},
  pages={22825--22855},
  year={2023},
  organization={PMLR}
}

@inproceedings{lightman2023let,
  title={Let's verify step by step},
  author={Lightman, Hunter and Kosaraju, Vineet and Burda, Yuri and Edwards, Harrison and Baker, Bowen and Lee, Teddy and Leike, Jan and Schulman, John and Sutskever, Ilya and Cobbe, Karl},
  booktitle={The Twelfth International Conference on Learning Representations},
  year={2023}
}

@article{liu2025video,
  title={Video-t1: Test-time scaling for video generation},
  author={Liu, Fangfu and Wang, Hanyang and Cai, Yimo and Zhang, Kaiyan and Zhan, Xiaohang and Duan, Yueqi},
  journal={arXiv preprint arXiv:2503.18942},
  year={2025}
}

@article{mi2025video,
  title={Video Generation Models Are Good Latent Reward Models},
  author={Mi, Xiaoyue and Yu, Wenqing and Lian, Jiesong and Jie, Shibo and Zhong, Ruizhe and Liu, Zijun and Zhang, Guozhen and Zhou, Zixiang and Xu, Zhiyong and Zhou, Yuan and others},
  journal={arXiv preprint arXiv:2511.21541},
  year={2025}
}

@article{wan2025wan,
  title={Wan: Open and advanced large-scale video generative models},
  author={Wan, Team and Wang, Ang and Ai, Baole and Wen, Bin and Mao, Chaojie and Xie, Chen-Wei and Chen, Di and Yu, Feiwu and Zhao, Haiming and Yang, Jianxiao and others},
  journal={arXiv preprint arXiv:2503.20314},
  year={2025}
}

@article{gao2025seedance,
  title={Seedance 1.0: Exploring the Boundaries of Video Generation Models},
  author={Gao, Yu and Guo, Haoyuan and Hoang, Tuyen and Huang, Weilin and Jiang, Lu and Kong, Fangyuan and Li, Huixia and Li, Jiashi and Li, Liang and Li, Xiaojie and others},
  journal={arXiv preprint arXiv:2506.09113},
  year={2025}
}

@misc{sora2025sora2,
  title={Sora 2},
  author={OpenAI},
  year={2025},
  howpublished={\url{https://openai.com/sora}},
}

@misc{veo2025veo3,
  title={Veo 3},
  author={Google DeepMind},
  year={2025},
  howpublished={\url{https://deepmind.google/models/veo}},
}

@misc{kling2025kling,
  title={Kling},
  author={Kuaishou Technology},
  year={2025},
  howpublished={\url{https://klingai.com/global}},
}

@article{luo2025sample,
  title={Sample By Step, Optimize By Chunk: Chunk-Level GRPO For Text-to-Image Generation},
  author={Luo, Yifu and Du, Penghui and Li, Bo and Du, Sinan and Zhang, Tiantian and Chang, Yongzhe and Wu, Kai and Gai, Kun and Wang, Xueqian},
  journal={arXiv preprint arXiv:2510.21583},
  year={2025}
}

@article{zhang2026grpo,
  title={E-GRPO: High Entropy Steps Drive Effective Reinforcement Learning for Flow Models},
  author={Zhang, Shengjun and Zhang, Zhang and Dai, Chensheng and Duan, Yueqi},
  journal={arXiv preprint arXiv:2601.00423},
  year={2026}
}

@article{he2025tempflow,
  title={Tempflow-grpo: When timing matters for grpo in flow models},
  author={He, Xiaoxuan and Fu, Siming and Zhao, Yuke and Li, Wanli and Yang, Jian and Yin, Dacheng and Rao, Fengyun and Zhang, Bo},
  journal={arXiv preprint arXiv:2508.04324},
  year={2025}
}

@article{ding2025treegrpo,
  title={TreeGRPO: Tree-Advantage GRPO for Online RL Post-Training of Diffusion Models},
  author={Ding, Zheng and Ye, Weirui},
  journal={arXiv preprint arXiv:2512.08153},
  year={2025}
}

@article{xu2023imagereward,
  title={Imagereward: Learning and evaluating human preferences for text-to-image generation},
  author={Xu, Jiazheng and Liu, Xiao and Wu, Yuchen and Tong, Yuxuan and Li, Qinkai and Ding, Ming and Tang, Jie and Dong, Yuxiao},
  journal={Advances in Neural Information Processing Systems},
  volume={36},
  pages={15903--15935},
  year={2023}
}

@article{li2025mixgrpo,
  title={Mixgrpo: Unlocking flow-based grpo efficiency with mixed ode-sde},
  author={Li, Junzhe and Cui, Yutao and Huang, Tao and Ma, Yinping and Fan, Chun and Yang, Miles and Zhong, Zhao},
  journal={arXiv preprint arXiv:2507.21802},
  year={2025}
}

@article{liu2025value,
  title={Value Gradient Guidance for Flow Matching Alignment},
  author={Liu, Zhen and Xiao, Tim Z and Domingo-Enrich, Carles and Liu, Weiyang and Zhang, Dinghuai},
  journal={arXiv preprint arXiv:2512.05116},
  year={2025}
}

@article{liu2025improving,
  title={Improving video generation with human feedback},
  author={Liu, Jie and Liu, Gongye and Liang, Jiajun and Yuan, Ziyang and Liu, Xiaokun and Zheng, Mingwu and Wu, Xiele and Wang, Qiulin and Xia, Menghan and Wang, Xintao and others},
  journal={arXiv preprint arXiv:2501.13918},
  year={2025}
}

@article{zheng2025vbench,
  title={Vbench-2.0: Advancing video generation benchmark suite for intrinsic faithfulness},
  author={Zheng, Dian and Huang, Ziqi and Liu, Hongbo and Zou, Kai and He, Yinan and Zhang, Fan and Gu, Lulu and Zhang, Yuanhan and He, Jingwen and Zheng, Wei-Shi and others},
  journal={arXiv preprint arXiv:2503.21755},
  year={2025}
}

@article{zheng2025diffusion,
  title={Diffusion transformers with representation autoencoders},
  author={Zheng, Boyang and Ma, Nanye and Tong, Shengbang and Xie, Saining},
  journal={arXiv preprint arXiv:2510.11690},
  year={2025}
}

@article{loshchilov2017decoupled,
  title={Decoupled weight decay regularization},
  author={Loshchilov, Ilya and Hutter, Frank},
  journal={arXiv preprint arXiv:1711.05101},
  year={2017}
}

@article{paszke2019pytorch,
  title={Pytorch: An imperative style, high-performance deep learning library},
  author={Paszke, Adam and Gross, Sam and Massa, Francisco and Lerer, Adam and Bradbury, James and Chanan, Gregory and Killeen, Trevor and Lin, Zeming and Gimelshein, Natalia and Antiga, Luca and others},
  journal={Advances in neural information processing systems},
  volume={32},
  year={2019}
}

@article{zhao2023pytorch,
  title={Pytorch fsdp: experiences on scaling fully sharded data parallel},
  author={Zhao, Yanli and Gu, Andrew and Varma, Rohan and Luo, Liang and Huang, Chien-Chin and Xu, Min and Wright, Less and Shojanazeri, Hamid and Ott, Myle and Shleifer, Sam and others},
  journal={arXiv preprint arXiv:2304.11277},
  year={2023}
}

@article{rafailov2023direct,
  title={Direct preference optimization: Your language model is secretly a reward model},
  author={Rafailov, Rafael and Sharma, Archit and Mitchell, Eric and Manning, Christopher D and Ermon, Stefano and Finn, Chelsea},
  journal={Advances in neural information processing systems},
  volume={36},
  pages={53728--53741},
  year={2023}
}

@article{zheng2023guided,
  title={Guided flows for generative modeling and decision making},
  author={Zheng, Qinqing and Le, Matt and Shaul, Neta and Lipman, Yaron and Grover, Aditya and Chen, Ricky TQ},
  journal={arXiv preprint arXiv:2311.13443},
  year={2023}
}
